\documentclass[10pt,twocolumn,letterpaper]{article}

\usepackage{btas}
\usepackage{times}
\usepackage{epsfig}
\usepackage{graphicx}
\usepackage{amsmath}
\usepackage{amssymb}
\usepackage[caption=false,font=footnotesize]{subfig}
\usepackage{tabularx}
\usepackage{booktabs}
\usepackage{verbatim}

\btasfinalcopy 

\usepackage{hyperref}

\hypersetup{
  colorlinks, linkcolor=red
}
\makeatletter
\g@addto@macro{\UrlBreaks}{\UrlOrds}
\makeatother
\newcolumntype{Y}{>{\centering\arraybackslash}X}
\DeclareRobustCommand{\eg} {\textit{e}.\textit{g}.\@\xspace}
\DeclareRobustCommand{\ie}{\textit{i}.\textit{e}.\@\xspace}
\DeclareRobustCommand{\etal}{\textit{et al.}\@\xspace}

\ifbtasfinal\fi
\begin{document}

\title{Face Recognition: Primates in the Wild\thanks{Earlier work on unconstrained human face recognition has been referred to as ``face recognition in the wild"~\cite{lfw}. In those studies, the term `wild' was used metaphorically. Here, we use the word `wild' literally.}}

\author{Debayan Deb$^1$, Susan Wiper$^2$, Alexandra H. Russo$^3$, Sixue Gong$^1$,\\Yichun Shi$^1$, Cori Tymoszek$^1$, and Anil K. Jain$^1$\\ \\
$^1$Department of Computer Science and Engineering, Michigan State University, East Lansing, MI, USA\\
$^2$University of Chester, UK, $^3$Conservation Biologist\\
{\small E-mail: $^1$\tt{\{debdebay, gongsixu, shiyichu, tymoszek, jain\}@cse.msu.edu},}\\
{\small$^2$\tt{s.wiper@chester.ac.uk}, $^3$\tt{alexandra.h.russo@gmail.com}}
}

\maketitle
\thispagestyle{empty}

\begin{abstract}
We present a new method of primate face recognition, and evaluate this method on several endangered primates, including golden monkeys, lemurs, and chimpanzees. The three datasets contain a total of 11,637 images of 280 individual primates from 14 species. Primate face recognition performance is evaluated using two existing state-of-the-art open-source systems, (i) FaceNet and (ii) SphereFace, (iii) a lemur face recognition system from literature, and (iv) our new convolutional neural network (CNN) architecture called PrimNet. Three recognition scenarios are considered: verification (1:1 comparison), and both open-set and closed-set identification (1:N search). We demonstrate that PrimNet outperforms all of the other systems in all three scenarios for all primate species tested. Finally, we implement an Android application of this recognition system to be assist primate researchers and conservationists in the wild for individual recognition of primates.
\end{abstract}

\section{Introduction}

In 2008, IUCN released a detailed report, \textit{Red List of Threatened Species}, which concluded that global diversity is severely threatened~\cite{iucn}. IUCN found that 22\% of all mammal species are `critically endangered', `endangered', or `vulnerable.' Primates, as an order of mammals, are particularly threatened, with around 60\% of all primate species and around 91\% of all lemur species threatened by extinction~\cite{primate_threat},~\cite{most_endangered}. Lemurs are native only to the island of Madagascar, where their forest habitat is being destroyed to make room for crops and feed illegal hardwood trade.~\cite{lemur_trouble}. Lemurs also fall prey to over-hunting as their meat is highly desired~\cite{iucn}. Similarly, the endangered golden monkey has endured extensive habitat loss and are now only found in a few national parks in Africa~\cite{chapman}. Intervention is necessary to halt and reverse these population declines of endangered primates, and one such intervention lies in individualization of these animals through automated facial recognition. Improved recognition and tracking will benefit the long-term health and stability of these species in a number of ways by (i) enabling more efficient longitudinal study, (ii) eliminating harmful effects of traditional tracking methods, and (iii) combating illegal trafficking and trade. This study proposes a non-invasive method of automatic facial recognition for primates which will be shown to be just as effective for golden monkeys, chimpanzees and indeed, we believe, any primate.

\begin{figure}[!t]
  \centering
  \subfloat[]{\includegraphics[height=1.2in]{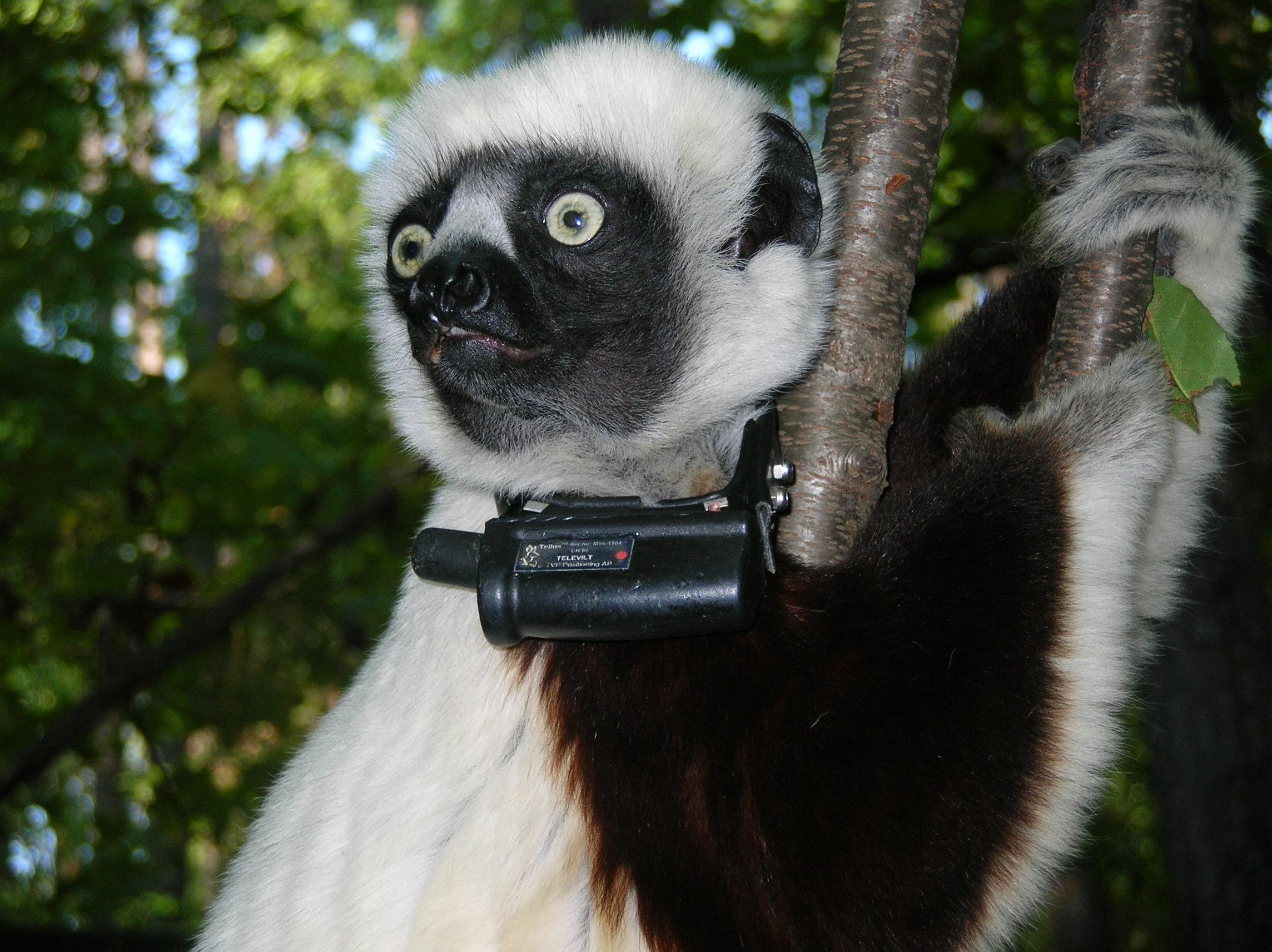}}\hfil
  \subfloat[]{\includegraphics[height=1.2in]{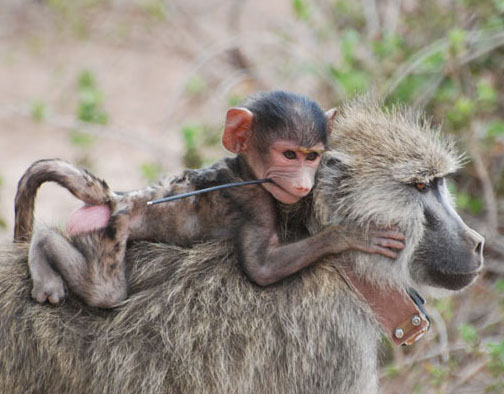}}
 \caption{Endangered Primates. (a) A lemur tagged and collared for tracking at Duke University Lemur Center~\cite{lemur_collared}. (b) A female savannah baboon wearing a GPS collar used for mammal tracking study~\cite{mammal_tracking}.}
  \label{fig:tag}
\end{figure}

Recognition of animals in the wild is critical for understanding the evolutionary processes that guide biodiversity. Researchers must reliably recognize each individual animal in order to observe that animal's variation within a population. Unique appearance-based cues, such as body size, presence of scars and marks, and coloring, are often used for interim studies~\cite{knowledge2}~\cite{knowledge3}, but these attributes are subjective and vary over time. Therefore, they are unreliable in longitudinal studies, which are necessary for the study of long-term population health and behavior, group dynamics, and the heritability and effects of traits~\cite{longitudinal}.

Biologists and anthropologists have started to adopt more objective and rigorous tracking methods, such as collars or tags (Figure~\ref{fig:tag}). While these approaches have been successfully used in several long-term in the wild primate studies~\cite{titi},~\cite{lemur1},~\cite{lemur4}, they are problematic in a number of ways. First, the devices can be expensive (\$400-\$4,000 per animal~\cite{collar_cost}) and time-consuming to apply. Second, tagging requires capture of the animal, which has demonstrably negative effects - it can disrupt social behavior~\cite{life1}, and cause intense stress~\cite{colobus}, injury~\cite{injury}, and even death~\cite{life2}. For the above reasons, the ethics of these methods have come under question~\cite{tagging_issues1},~\cite{tagging_issues2}. In contrast, automatic facial recognition is a promising method to accurately identify individuals with minimal risk to these already threatened species.

\begin{figure}[!t]
    \centering
  \subfloat[]{\includegraphics[width=0.48\linewidth]{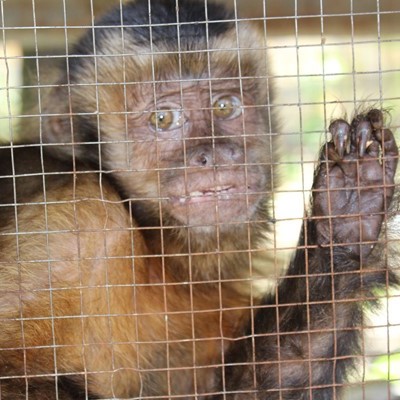}}\hfil
  \subfloat[]{\includegraphics[width=0.48\linewidth]{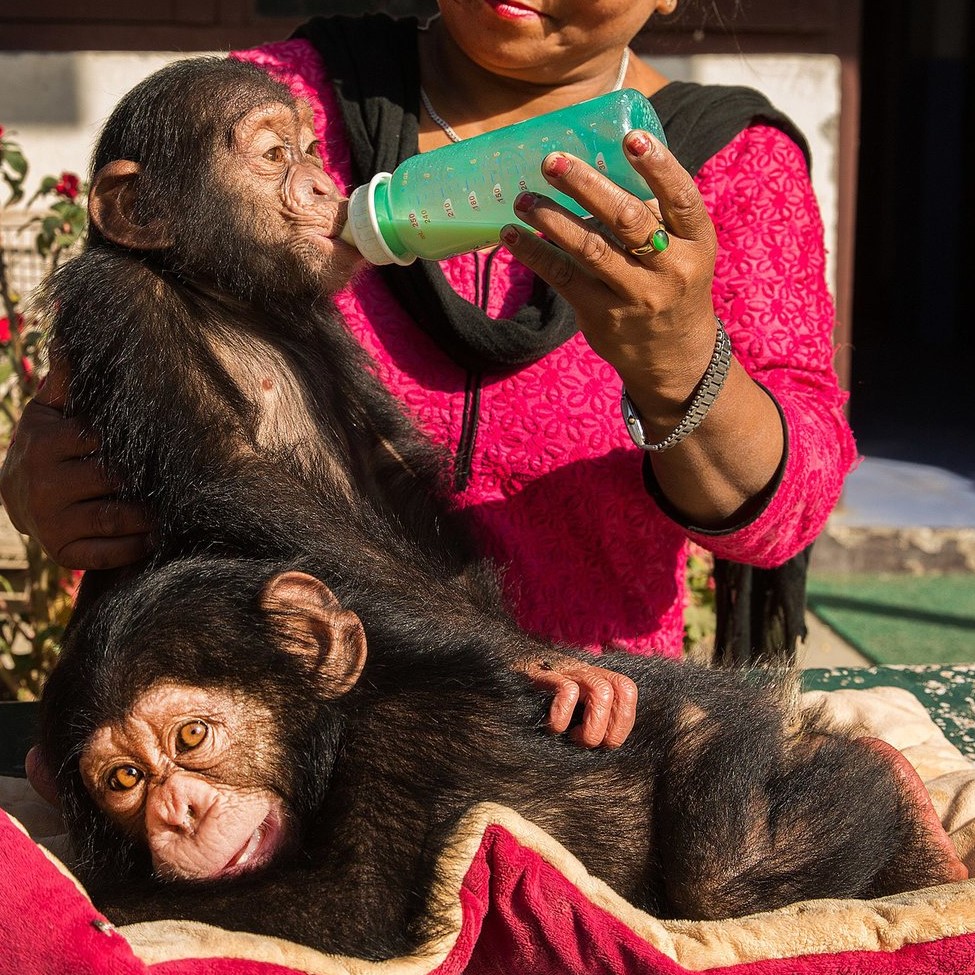}}
  \caption{Trafficking in primates. (a) A caged capuchin monkey in Peru.~\cite{capuchin}. (b) Two chimpanzees rescued from a smuggling operation in Kathmandu, Nepal~\cite{chimps_rescued}. }
   \label{fig:trafficking_1}
\end{figure}

\begin{figure}[!t]
  \centering
  \includegraphics[width=0.9\linewidth]{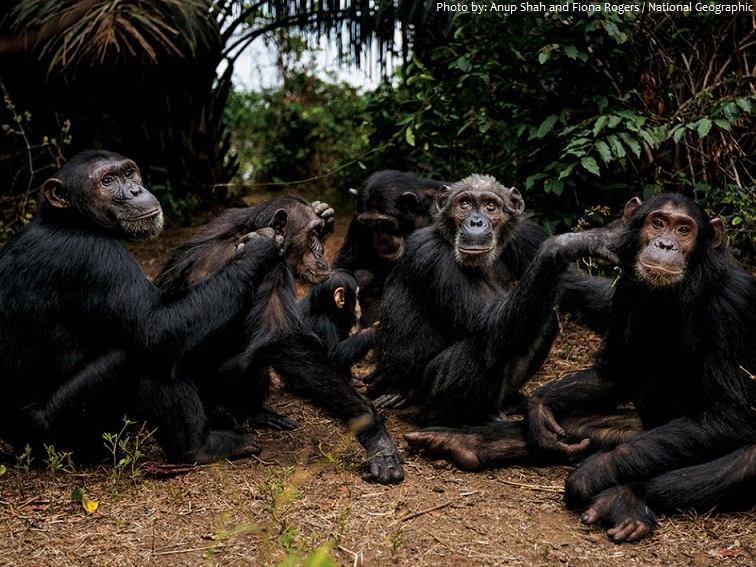}\hfil
  \label{chimp_party}
  \caption{Group of chimpanzees partying in the wild~\cite{party}.}
\end{figure}

A third opportunity to safeguard these endangered primate species lies in the growing problem of trafficking. Primate trafficking is a booming business in which these animals are captured from the wild for shipment around the globe (Figure~\ref{fig:trafficking_1}). In the case of great apes, for example, it is estimated that 22,218 individuals were lost between 2005 and 2011 due to illegal trade~\cite{stolen_apes}. In contrast, only 27 arrests were made in connection with such trade, indicating that little has been done to solve the problem~\cite{stolen_apes}. There is evidence that this illegal trade of great apes has been increasing since 2011~\cite{daniel}. If a captured individual can be identified, this will provide information about the animal's origin, and may provide insight into their capture.

There is an urgent need for a non-invasive, reliable method of identifying individuals that can be easily employed in the field. Kuhl \etal proposed animal biometrics as a potential solution~\cite{kuhl}. Computer-aided identification of individuals has been shown to be promising for wild animal populations such as cheetahs~\cite{cheetah}, tigers~\cite{tiger}, giraffes~\cite{giraffe}, zebras~\cite{zebra}, and penguins~\cite{penguin}. Primates are particularly promising as facial recognition targets because humans belong to the biological group known as primates. Humans are particularly close to great apes, as both are grouped together in one of the major groups of the primate evolutionary tree. Since primate facial structure is similar to that of humans (forward-facing eye sockets, small or absent snout), we expect that established human facial recognition techniques will generalize well to primate faces. Indeed, Freytag \etal worked on automatic individualization of chimpanzees in the wild~\cite{freytag} using Convolutional Neural Networks (CNN) and achieved ~92\% identification accuracy on a dataset containing 2,109 face images of 24 chimpanzees. Crouse~\etal proposed a face recognition system for lemurs (LemurFaceID) using simple LBP features~\cite{crouse}. LemurFaceID focused on individual identification of 80 red-bellied lemurs from Duke Lemur Center, and correctly identified individuals at Rank-1 accuracy of $98.7\% \pm 1.81\%$ (using 2-query image fusion). LemurFaceID solely focused on the identification scenario (1:N comparison). However, for automatic primate face recognition, validating whether a set of photographs belong to the same individual (1:1 comparison) is equally important. 

The aforementioned studies have not been implemented in a manner where a human operator can quickly perform identification in the wild using, say, a mobile app. To that effect, researchers from a Cornell lab developed an application, Merlin Bird ID~\cite{bird_id}, that can identify the species of birds, though it does not support individual identification. In this paper, we propose a non-invasive, rapid, and robust method of automatic primate individual identification which has been implemented as an Android smartphone application for rapid deployment and use.

Concisely, the contributions of the paper are as follows:
\begin{enumerate}
    \item Evaluated lemur individual identification performance of state-of-the-art and open-source human face recognition systems, FaceNet\footnote{\url{https://github.com/davidsandberg/facenet}}~\cite{facenet} and SphereFace\footnote{\url{https://github.com/wy1iu/sphereface}}~\cite{liu2017sphereface}\footnote{FaceNet and SphereFace achieve 99.65\% and 99.42\% accuracy on LFW dataset using the standard LFW protocol~\cite{lfw}, respectively.} on a dataset of 3,000 face images of 129 lemurs. SphereFace achieved an identification performance of 92.45\% at Rank-1.
    \item Proposed a new CNN architecture (PrimNet) suitable for small datasets available for primate faces that is implemented on a mobile phone. PrimNet achieves 93.75\% lemur individual identification accuracy at Rank-1.
    \item Demonstrated the generalization of PrimNet to other primates, such as chimpanzees and golden monkeys. PrimNet achieves Rank-1 accuracies of 90.26\% and 75.82\% for golden monkeys and chimpanzees, respectively.
    \item Implemented an Android app that can be used by primate researchers and conservationists in the wild for recognition (both 1:1 and 1:N) and tracking of primates.
    \item We plan to publicly open-source both the LemurFace and GoldenMonkeyFace datasets in order for other researchers to push the state-of-the-art in primate face recognition. In addition, the software for PrimNet, along with the mobile app, will also be open-sourced.
\end{enumerate}

\begin{figure}[!t]
  \centering
  \captionsetup[subfigure]{labelformat=empty, justification=centering}
  \subfloat[\textit{Eulemur coronatus}][\textit{Eulemur coronatus} \\ Crowned lemur]{\includegraphics[width=0.32\linewidth]{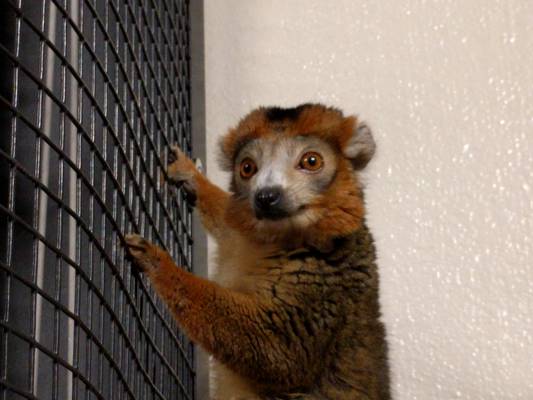}}\hfil
\subfloat[\textit{Propithecus coquereli}][\textit{Propithecus coquereli}\\ Coquerel's sifaka]{\includegraphics[width=0.32\linewidth]{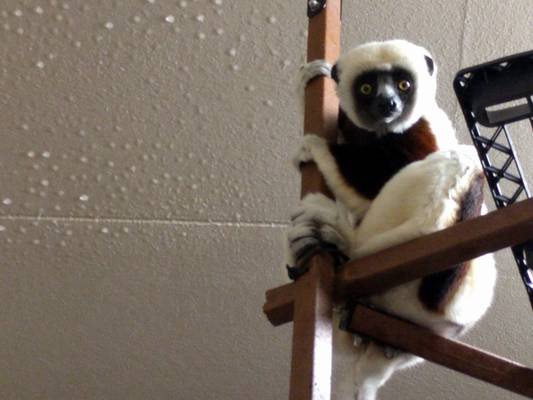}}\hfil
  \subfloat[\textit{Lemur catta}][\textit{Lemur catta}\\Ring-tailed lemur]{\includegraphics[width=0.32\linewidth]{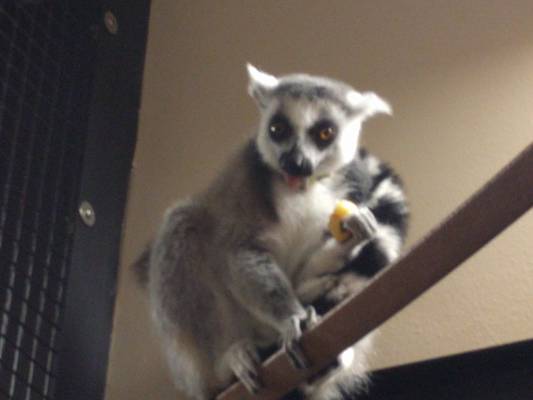}}\vfil
  \subfloat[\textit{Varecia variegata}][\textit{Varecia variegata}\\B/W ruffed lemur]{\includegraphics[width=0.32\linewidth]{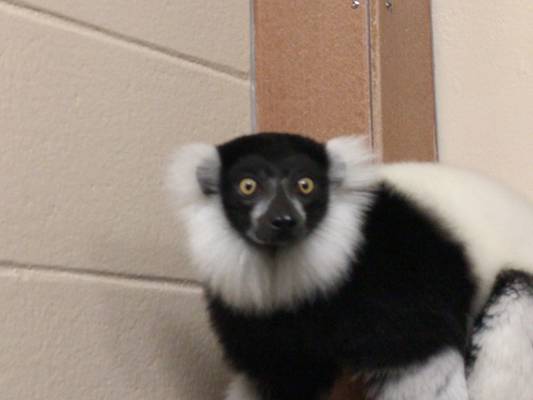}}\hfil
  \subfloat[\textit{Eulemur collaris}][\textit{Eulemur collaris}\\Collared brown lemur]{\includegraphics[width=0.32\linewidth]{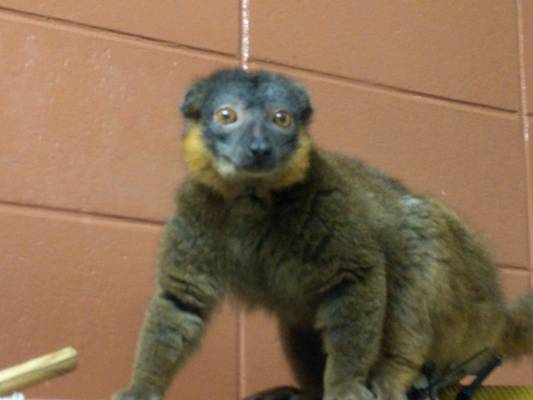}}\hfil
  \subfloat[\textit{Eulemur mongoz}][\textit{Eulemur mongoz}\\Mongoose lemur]{\includegraphics[width=0.32\linewidth]{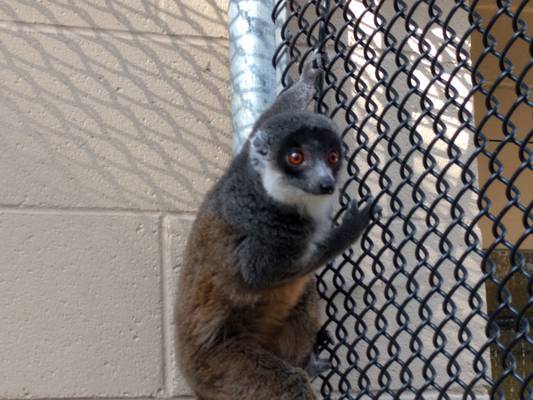}}\vfil
  \subfloat[\textit{Varecia rubra}][\textit{Varecia rubra}\\Red-ruffed lemur]{\centering\includegraphics[width=0.32\linewidth]{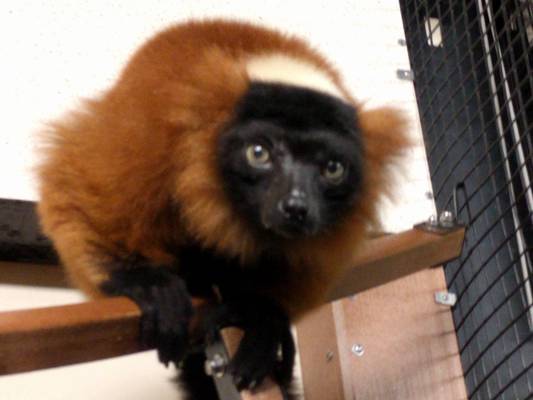}}\hfil
  \subfloat[\textit{Eulemur rubriventer}][\textit{Eulemur rubriventer}\\Red-bellied lemur]{\includegraphics[width=0.32\linewidth]{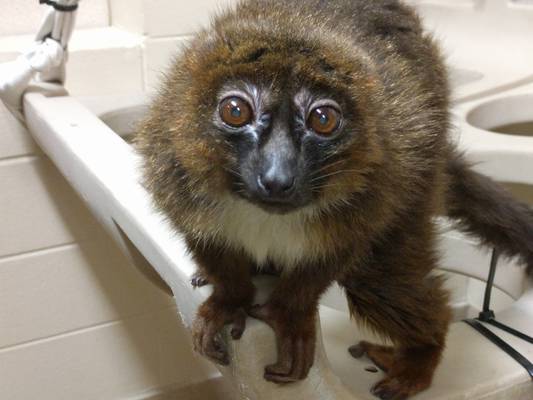}}\hfil
  \subfloat[\textit{Eulemur flavifrons}][\textit{Eulemur flavifrons}\\Blue-eyed black lemur]{\includegraphics[width=0.32\linewidth]{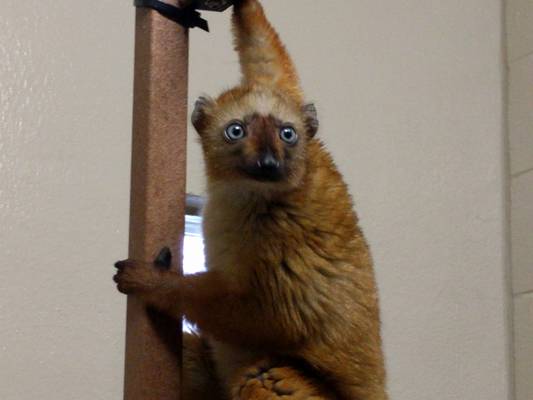}}
  \vspace{0.5em}
  \caption{Images of 9 out of 12 different lemur species in our dataset.}
  \label{fig:lemur_species}
\end{figure}

\begin{figure}[!t]
    \subfloat[Adam]{
    \begin{minipage}{0.49\textwidth}
    \centering
    \includegraphics[width=0.32\linewidth]{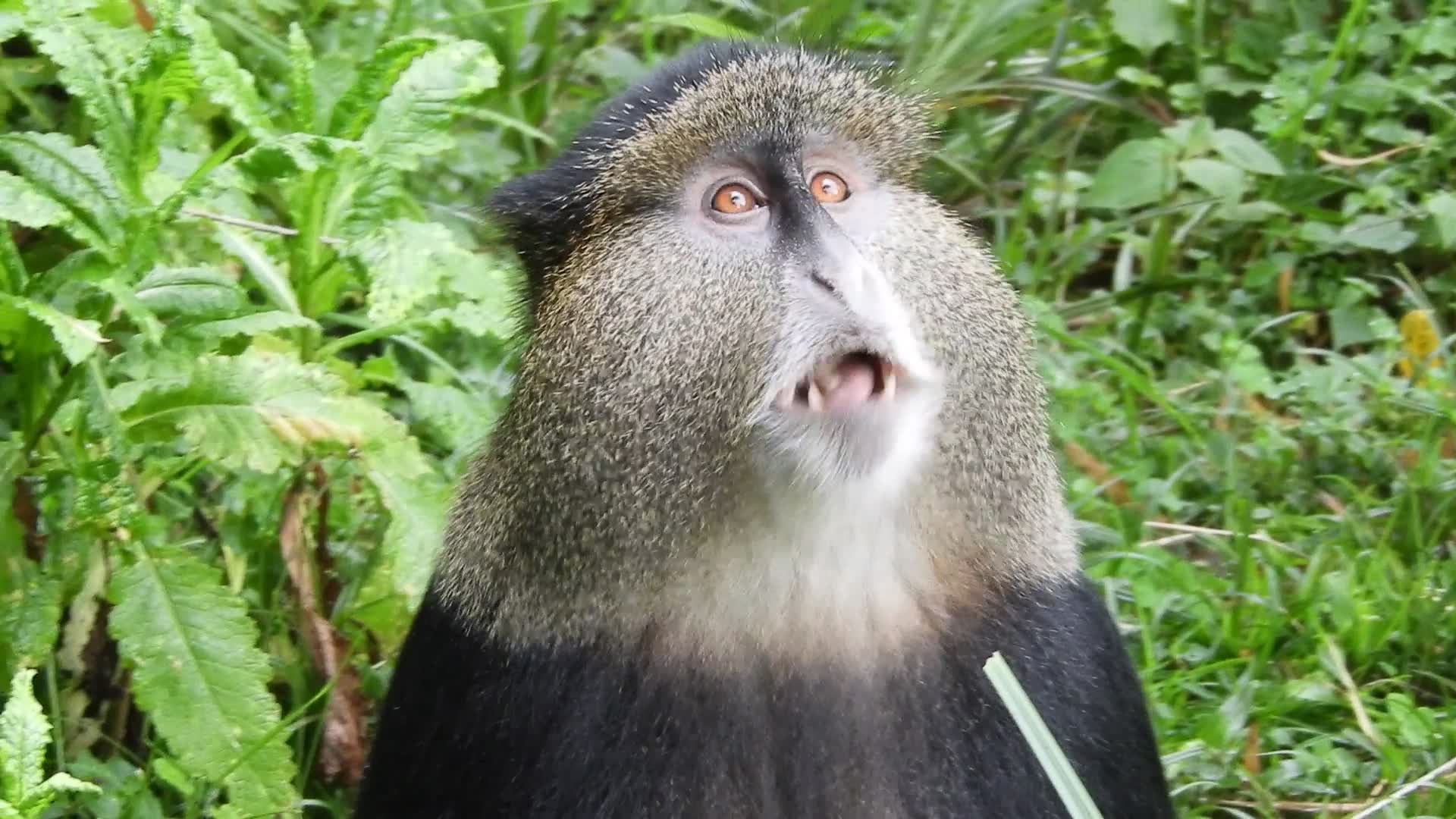}\hfil
    \includegraphics[width=0.32\linewidth]{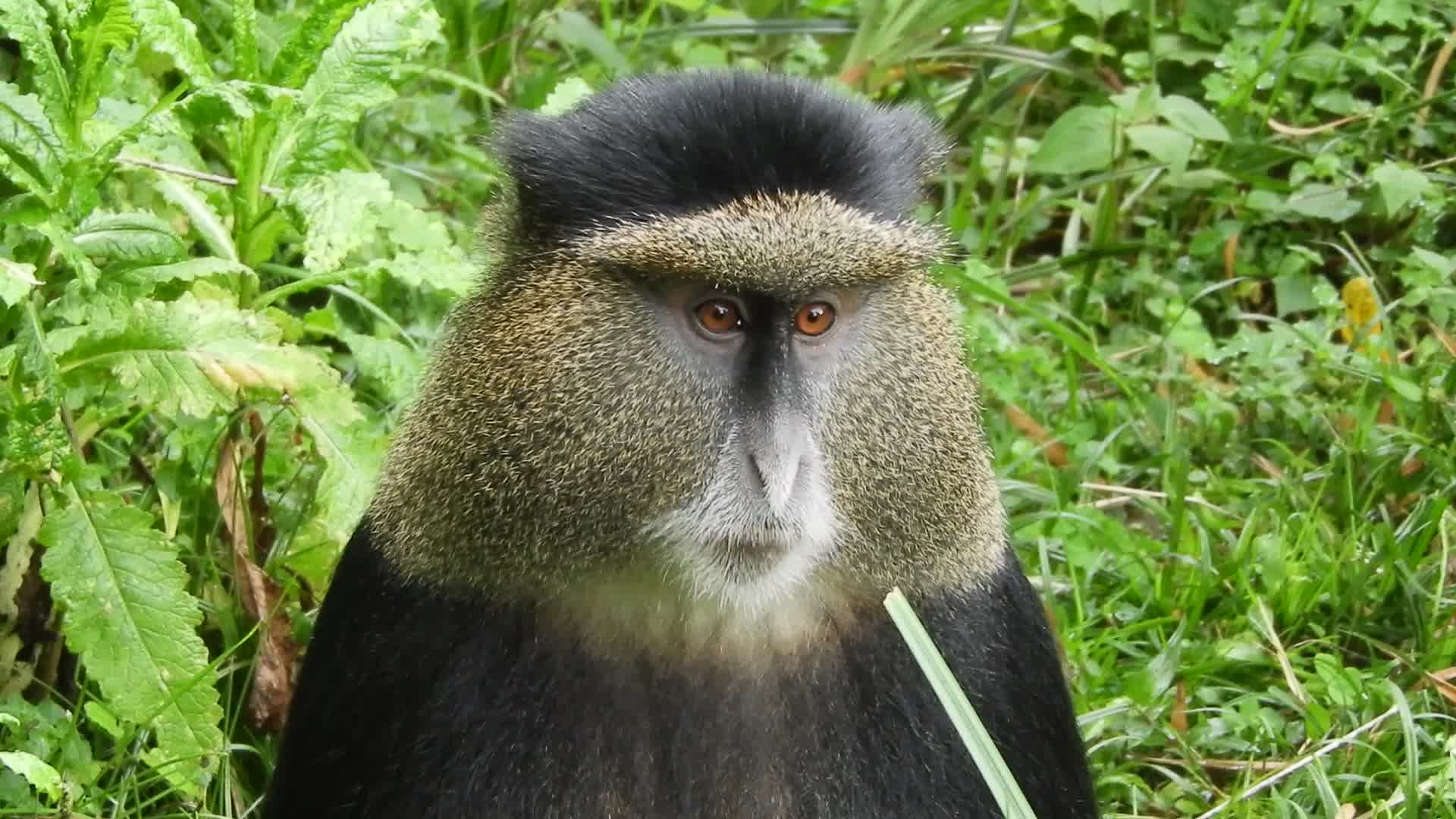}\hfil
    \includegraphics[width=0.32\linewidth]{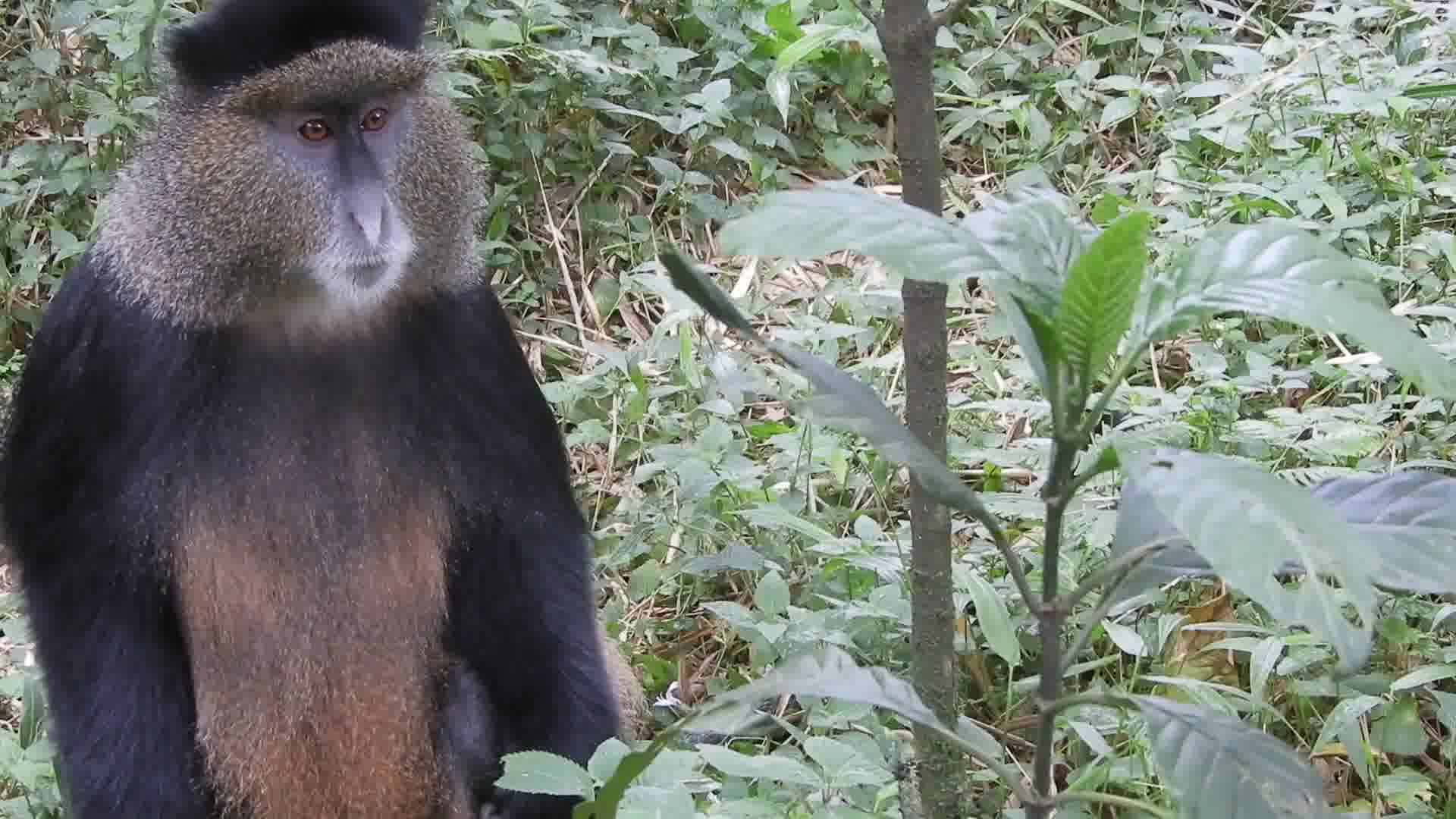}
    \end{minipage}}\hfil
    \subfloat[Dave]{
    \begin{minipage}{0.49\textwidth}
    \centering
    \includegraphics[width=0.32\linewidth]{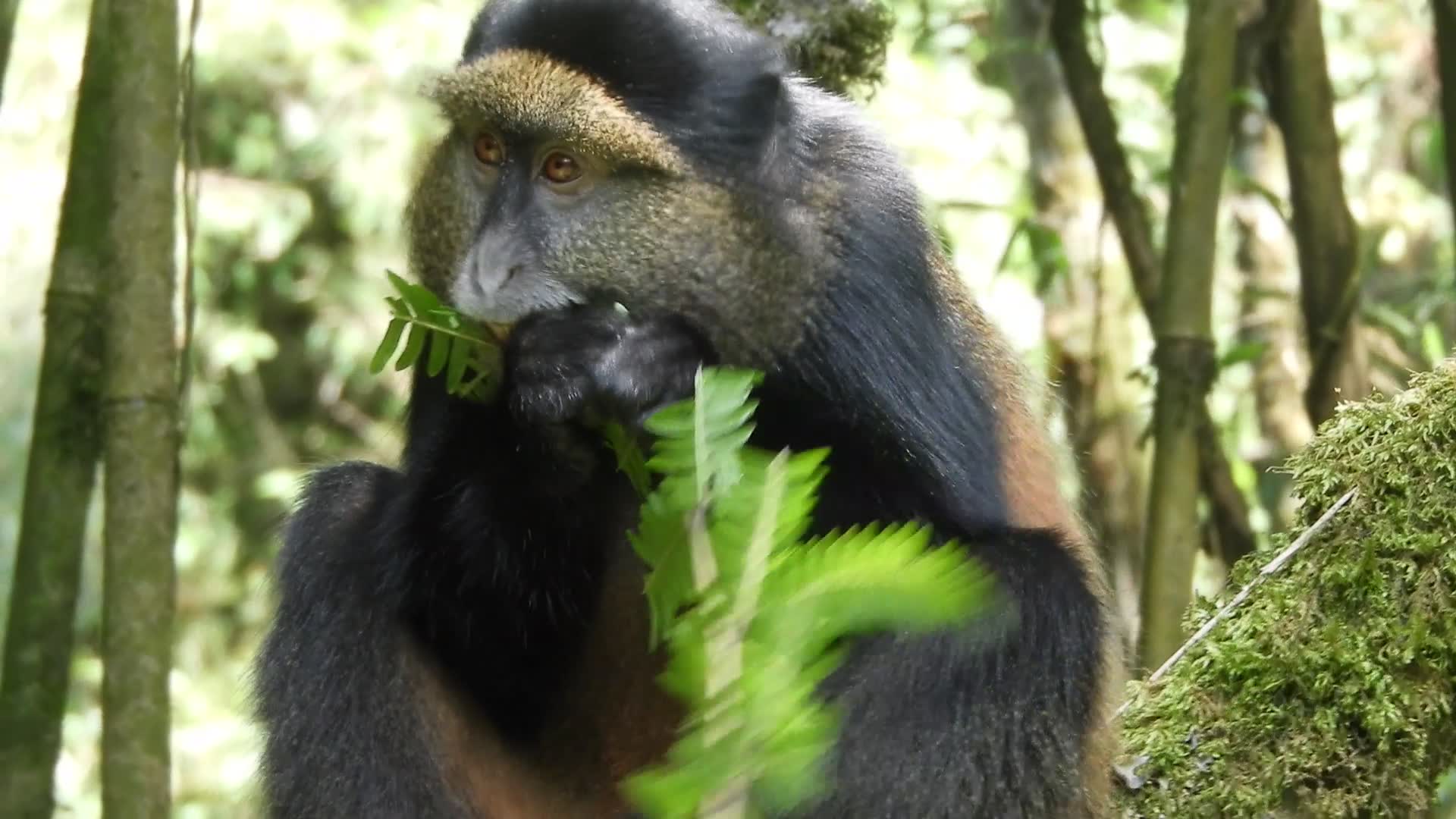}\hfil
      \includegraphics[width=0.32\linewidth]{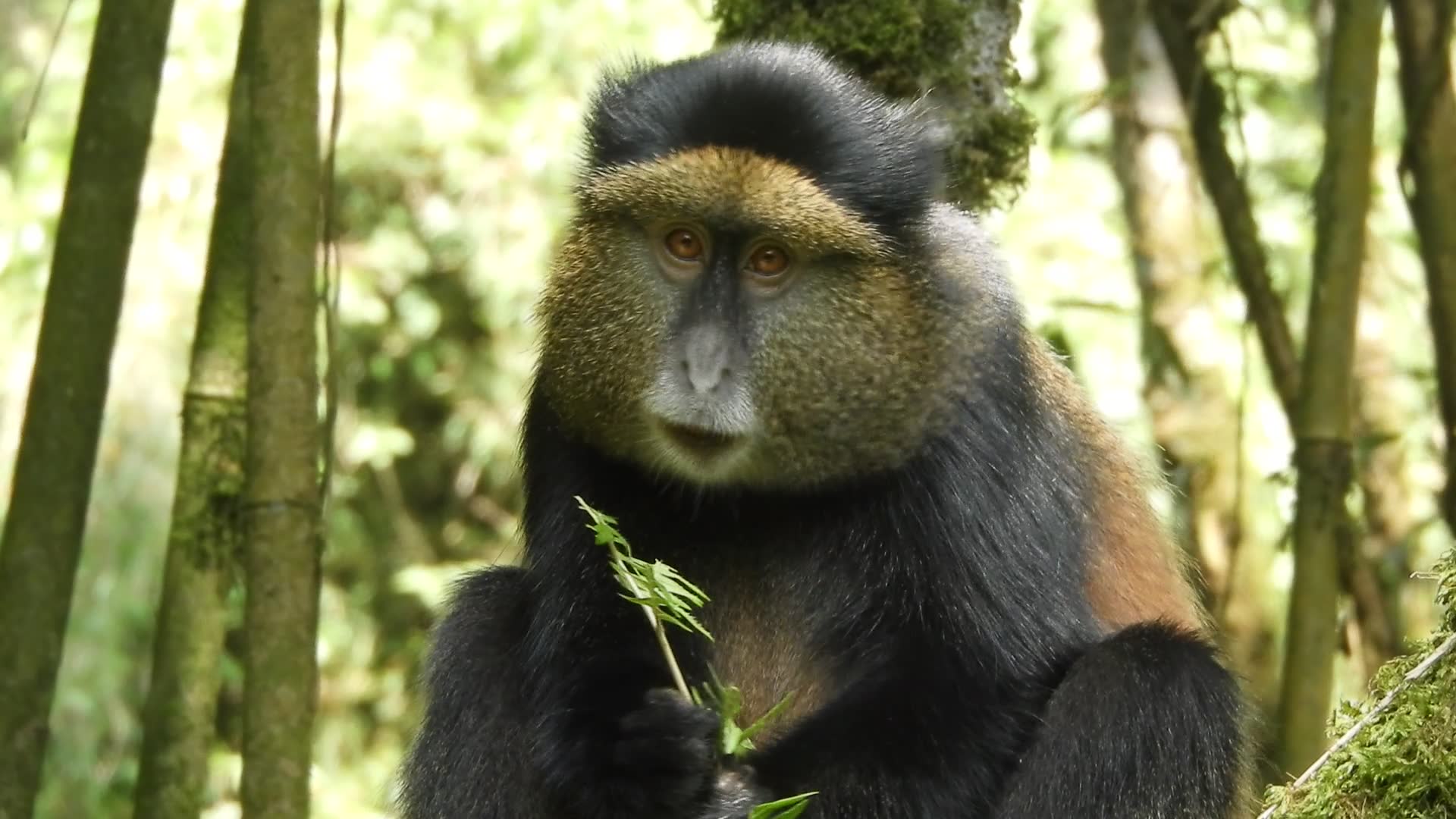}\hfil
      \includegraphics[width=0.32\linewidth]{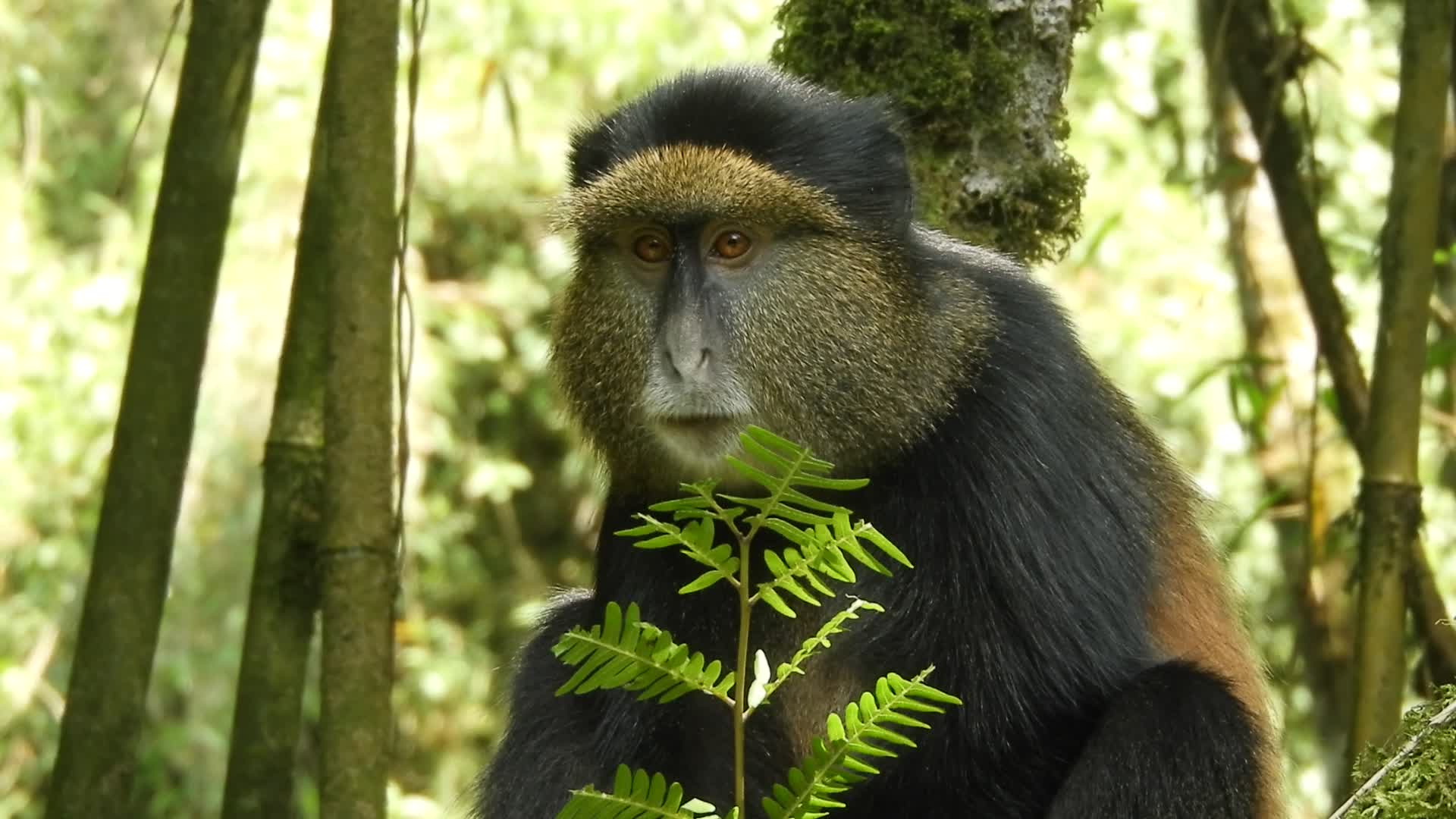}
    \end{minipage}}\vfil
    \subfloat[Duncan]{
    \begin{minipage}{0.49\textwidth}
    \centering
    \includegraphics[width=0.32\linewidth]{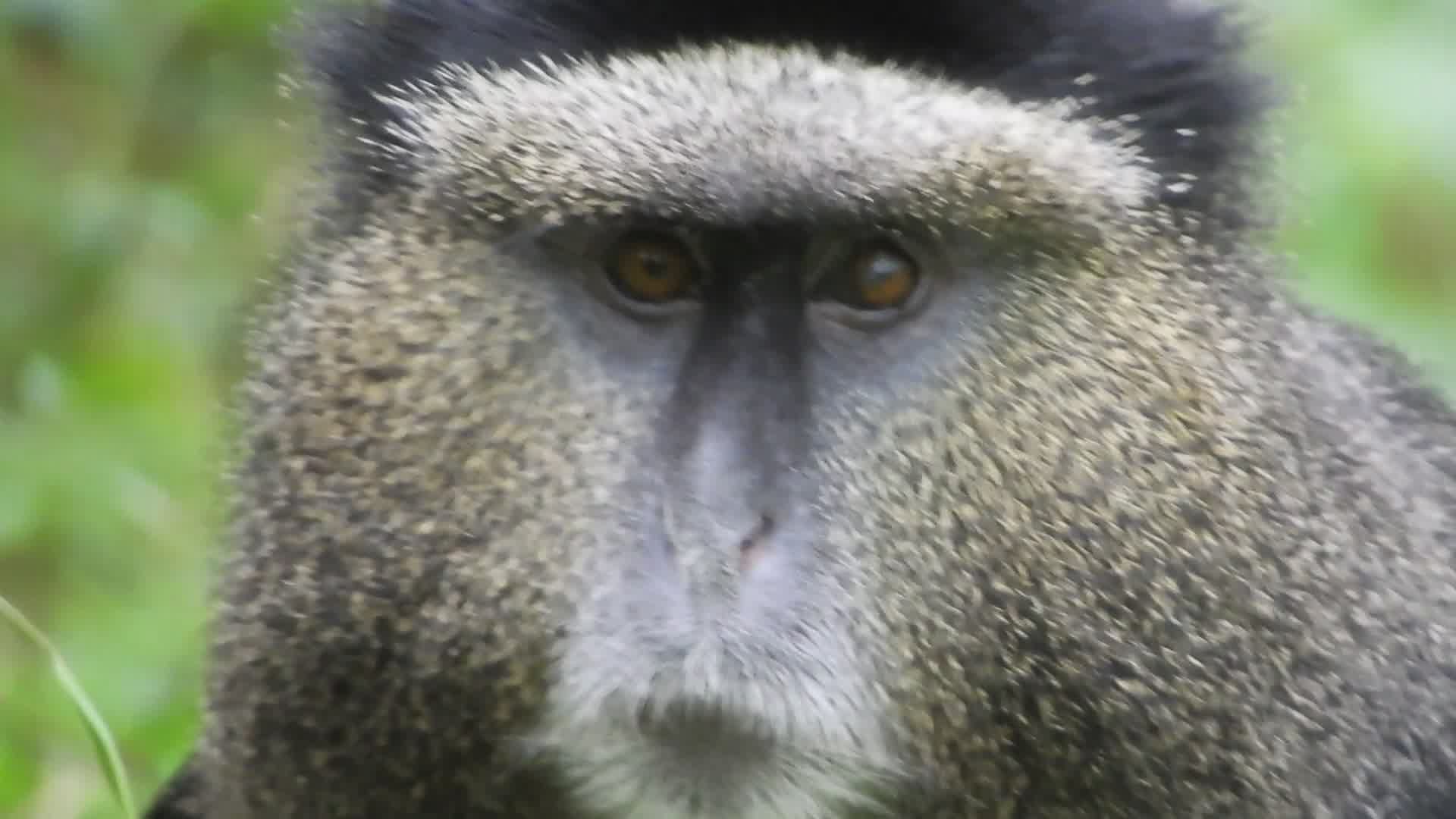}\hfil
  \includegraphics[width=0.32\linewidth]{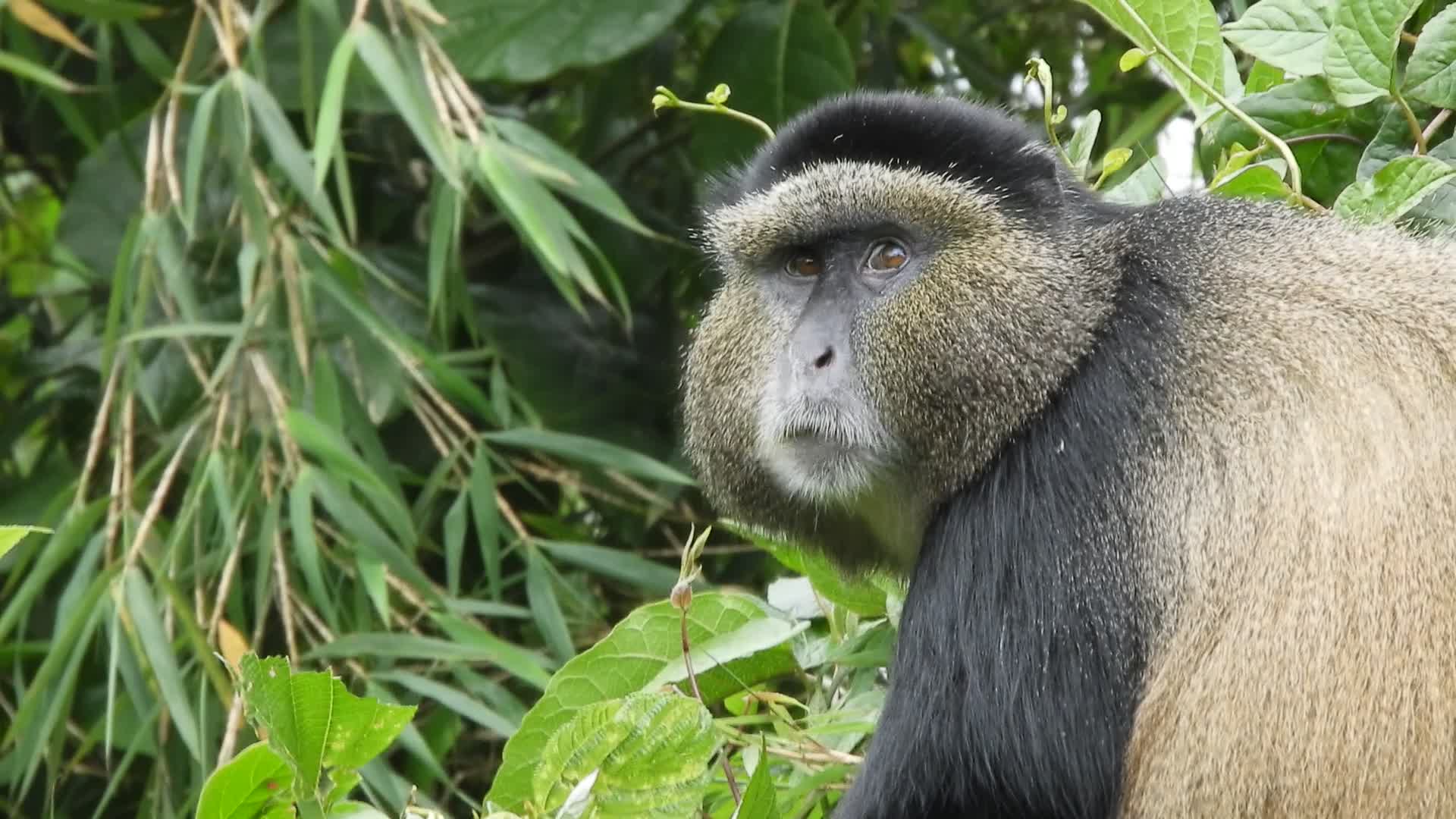}\hfil
  \includegraphics[width=0.32\linewidth]{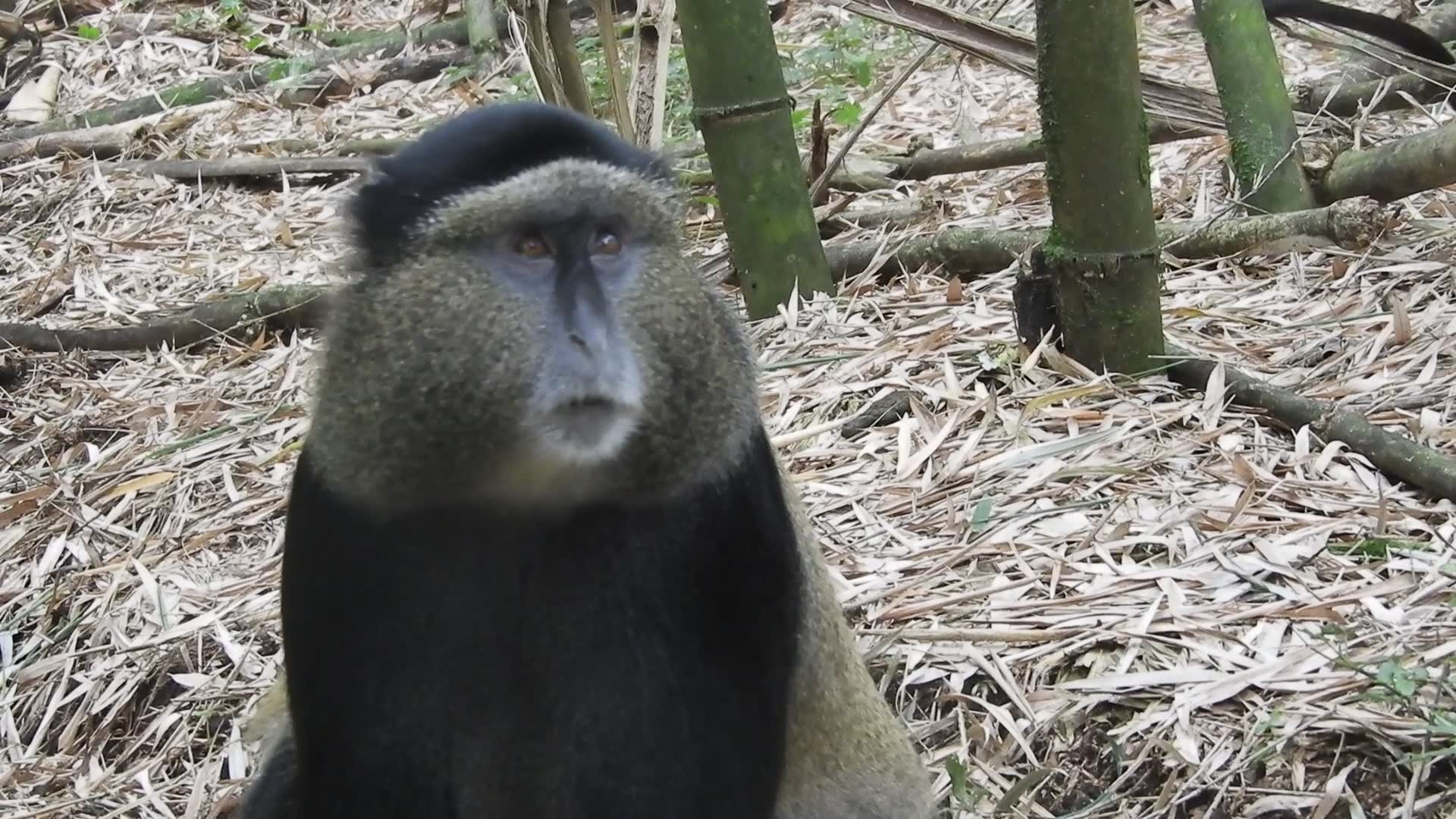}
    \end{minipage}}\hfil
    \subfloat[Ella]{
    \begin{minipage}{0.49\textwidth}
    \centering
    \includegraphics[width=0.32\linewidth]{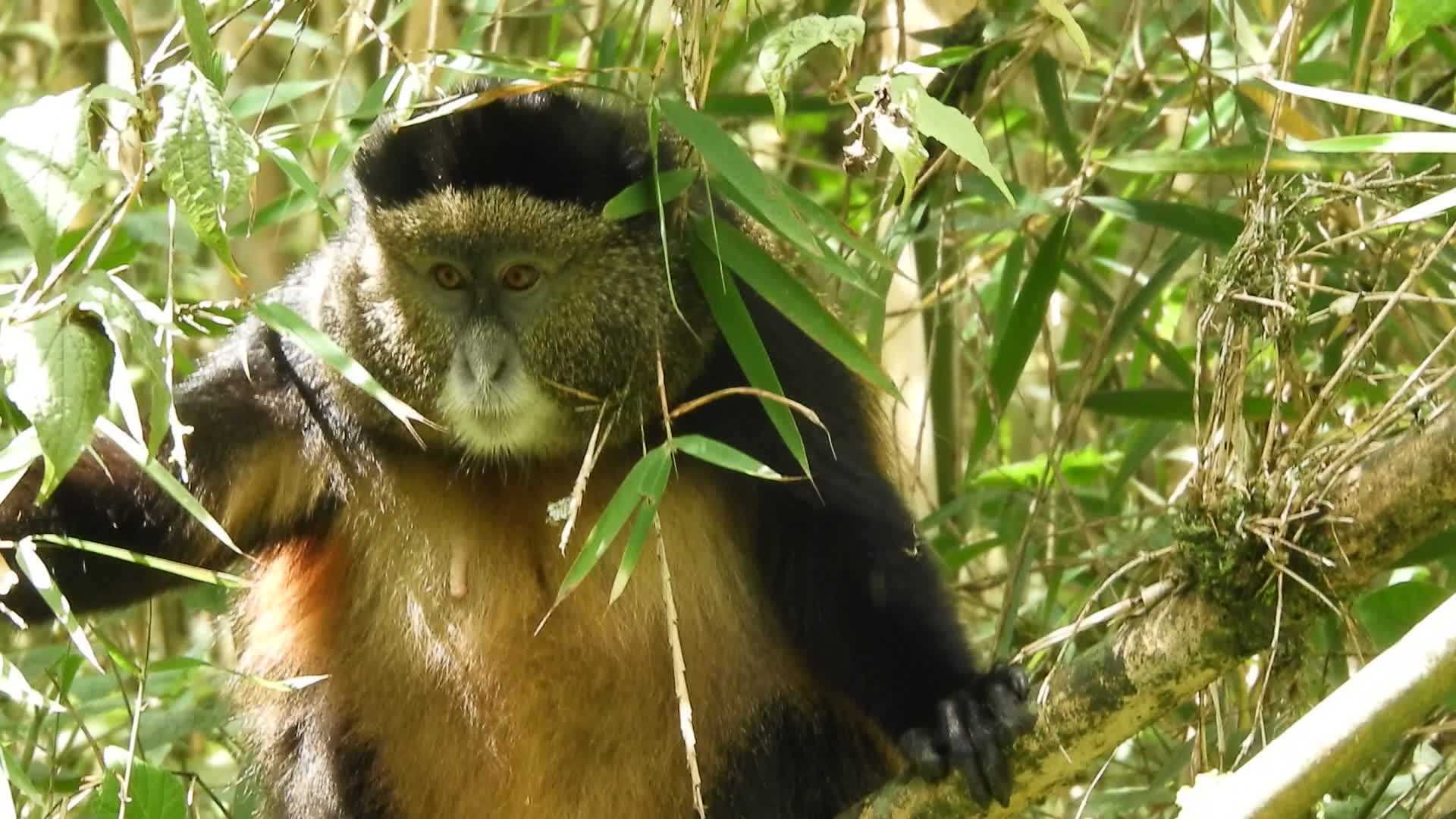}\hfil
      \includegraphics[width=0.32\linewidth]{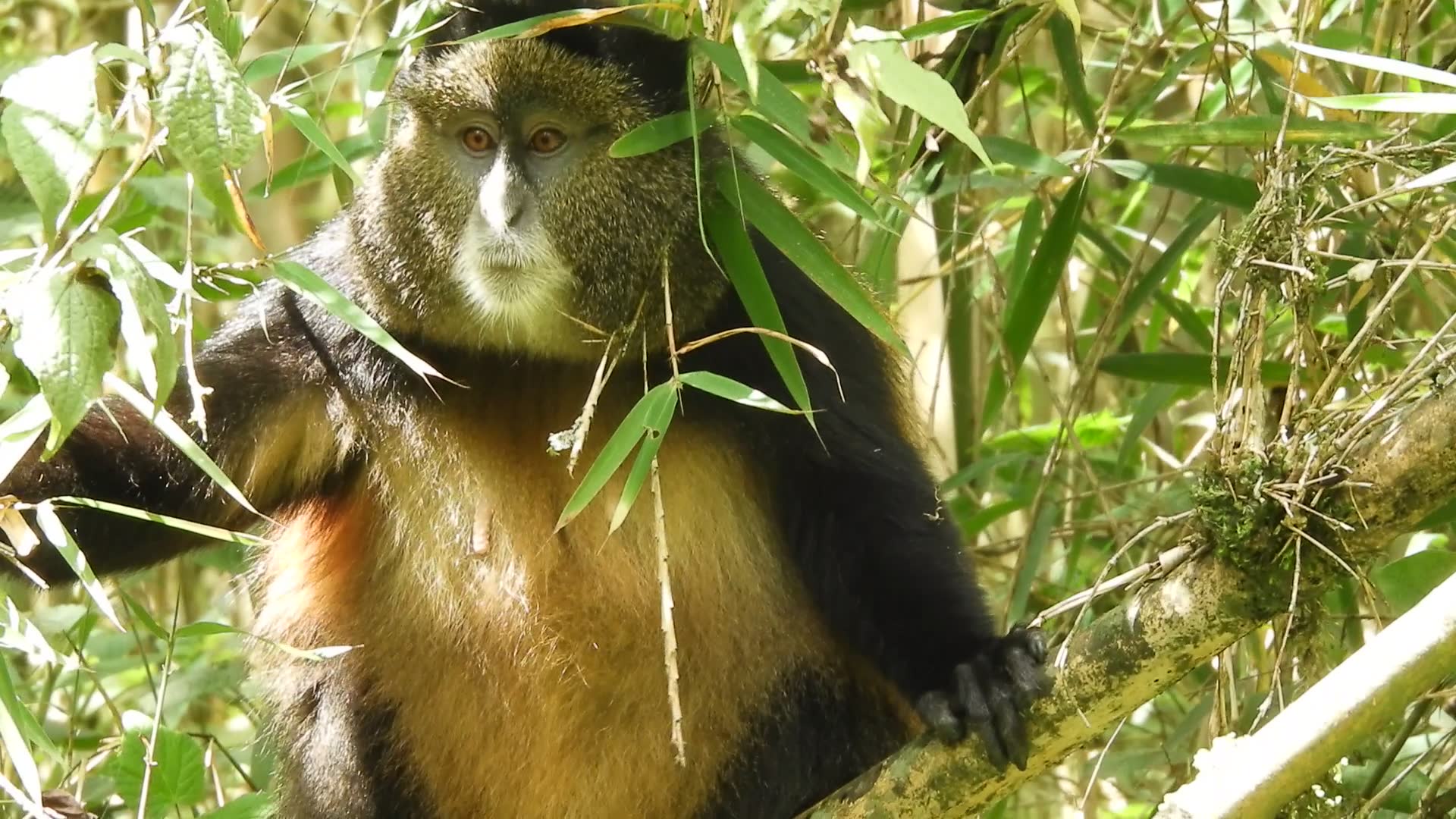}\hfil
      \includegraphics[width=0.32\linewidth]{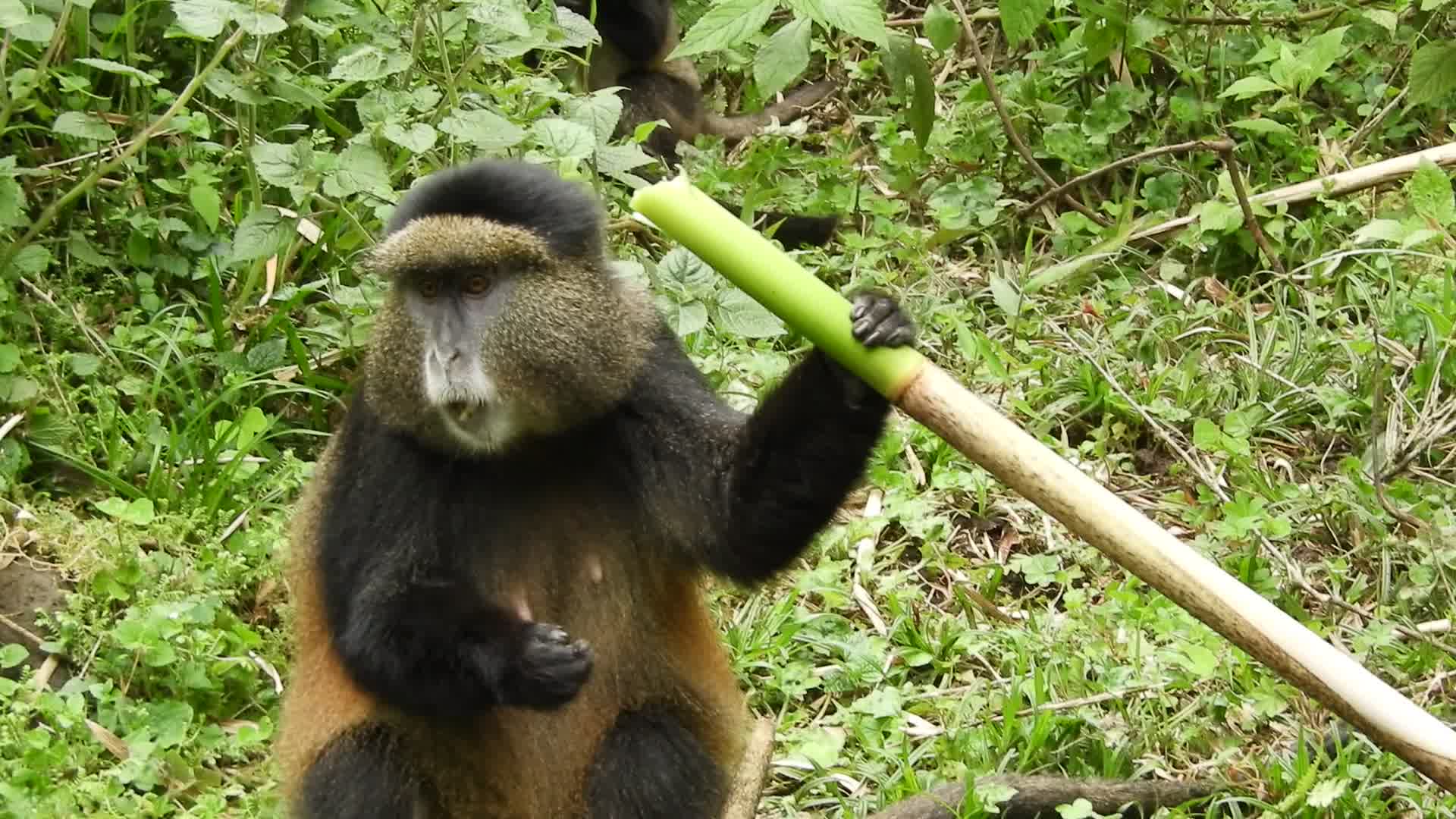}
    \end{minipage}}
  \caption{Images of four different golden monkeys in our dataset: \textit{Adam}, \textit{Dave}, \textit{Duncan}, and \textit{Ella}.}
  \label{fig:gm_samples}
\end{figure}

\begin{figure}[!t]
    \subfloat[Coco]{
    \begin{minipage}{0.49\textwidth}
    \centering
    \includegraphics[height=0.8in]{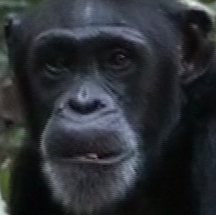}\hfil
    \includegraphics[height=0.8in]{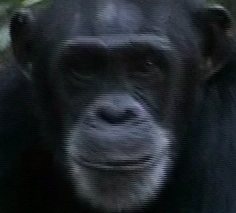}\hfil
    \includegraphics[height=0.8in]{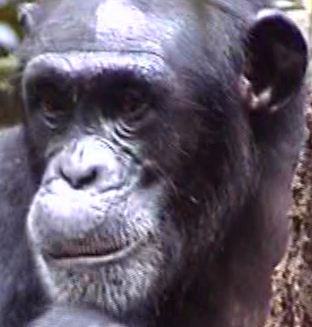}
    \end{minipage}}\hfil
    \subfloat[Fredy]{
    \begin{minipage}{0.49\textwidth}
    \centering
    \includegraphics[height=0.8in]{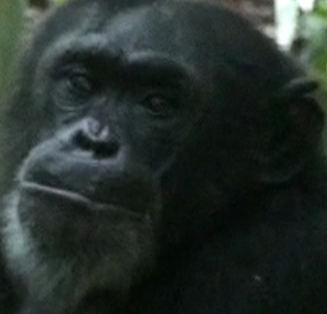}\hfil
      \includegraphics[height=0.8in]{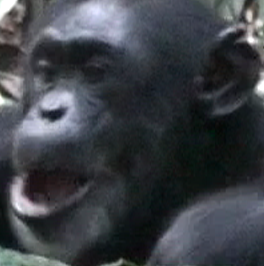}\hfil
      \includegraphics[height=0.8in]{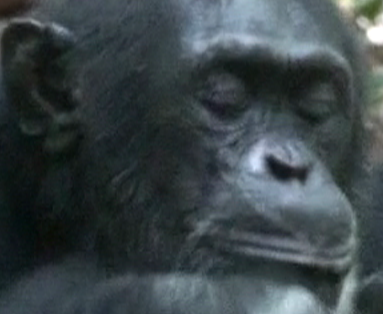}
    \end{minipage}}\vfil
    \subfloat[Oscar]{
    \begin{minipage}{0.49\textwidth}
    \centering
    \includegraphics[height=0.8in]{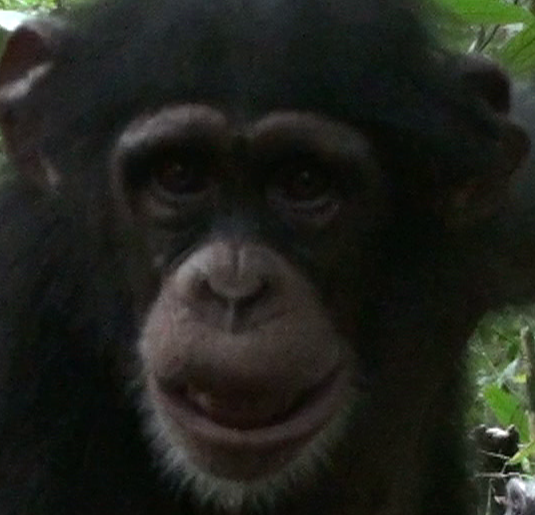}\hfil
  \includegraphics[height=0.8in]{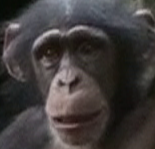}\hfil
  \includegraphics[height=0.8in]{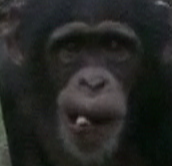}
    \end{minipage}}
  \caption{Images of three different chimpanzees in our dataset: \textit{Coco}, \textit{Fredy}, and \textit{Oscar}.}
  \label{fig:chimp_samples}
\end{figure}

\section{Dataset}
\label{dataset}
For our experiments, we acquired datasets of three different primates in the wild: lemurs, golden monkeys, and chimpanzees. In this paper, we refer to the datasets as \textit{LemurFace}, \textit{GoldenMonkeyFace}\footnote{Both LemurFace and GoldenMonkeyFace datasets are available for download at \url{https://github.com/ronny3050/PrimateFaceRecognition}.}, and \textit{ChimpFace}\footnote{\url{https://github.com/cvjena/chimpanzee_faces}}~\cite{loos},~\cite{freytag}, respectively.

\subsection{LemurFace Dataset}
\begin{figure}[!t]
  \centering
  \subfloat[Lemurs]{\includegraphics[width=0.49\linewidth]{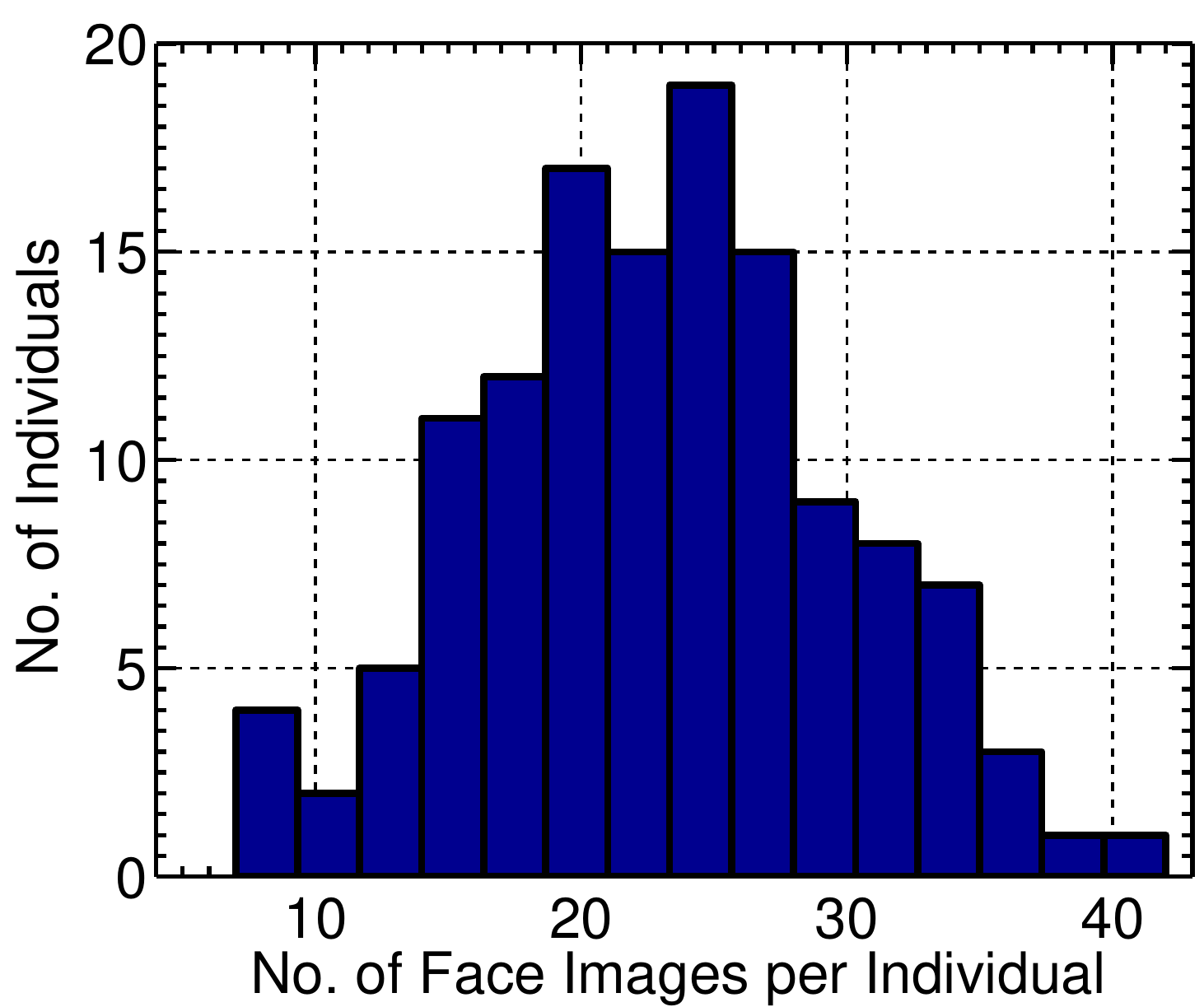}\label{fig:lemur_imgsPerInd}}\hfil
  \subfloat[Golden Monkeys]{\includegraphics[width=0.49\linewidth]{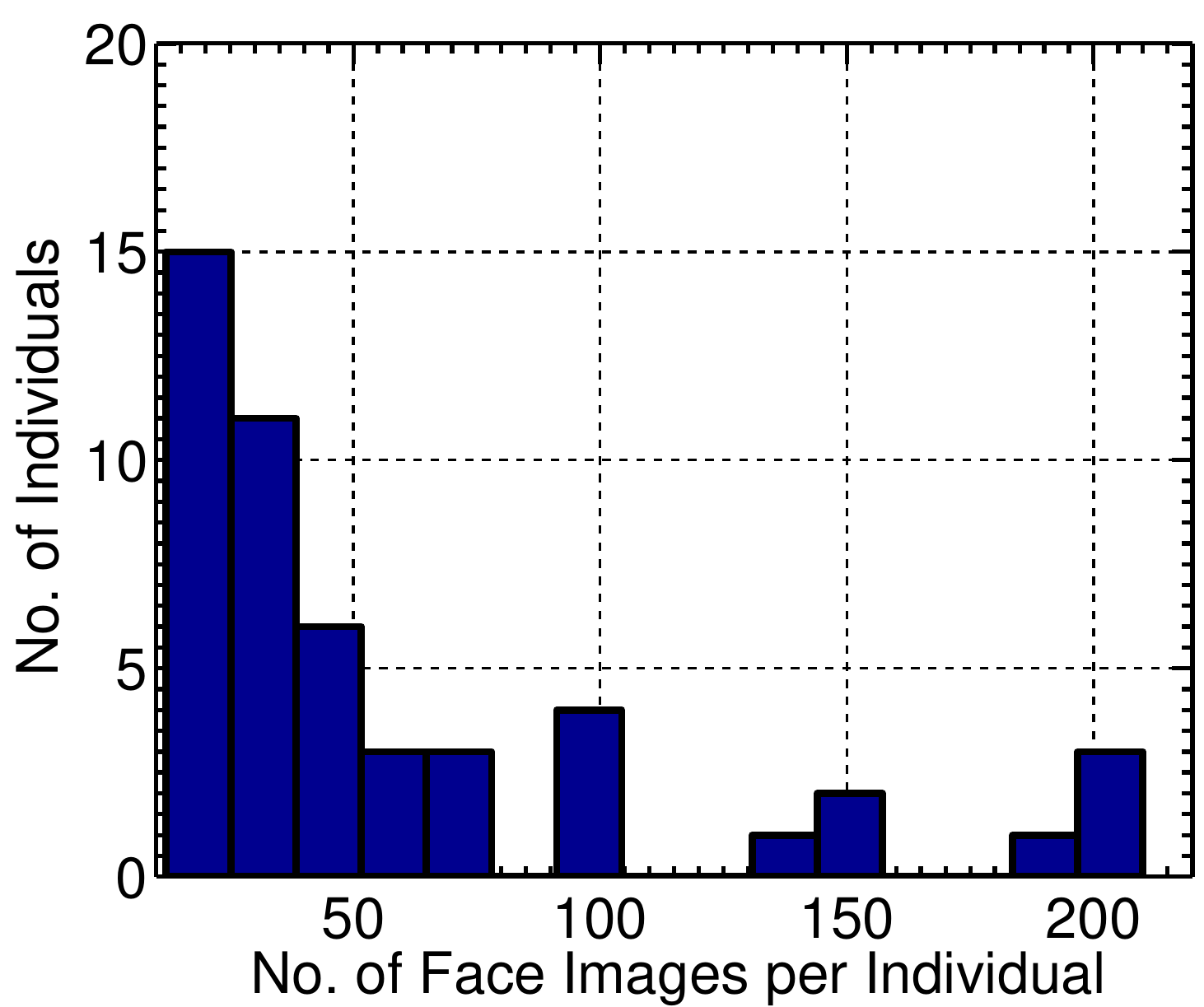}\label{fig:gm_imgsPerInd}}\vfil
  \subfloat[Chimpanzees]{\includegraphics[width=0.49\linewidth]{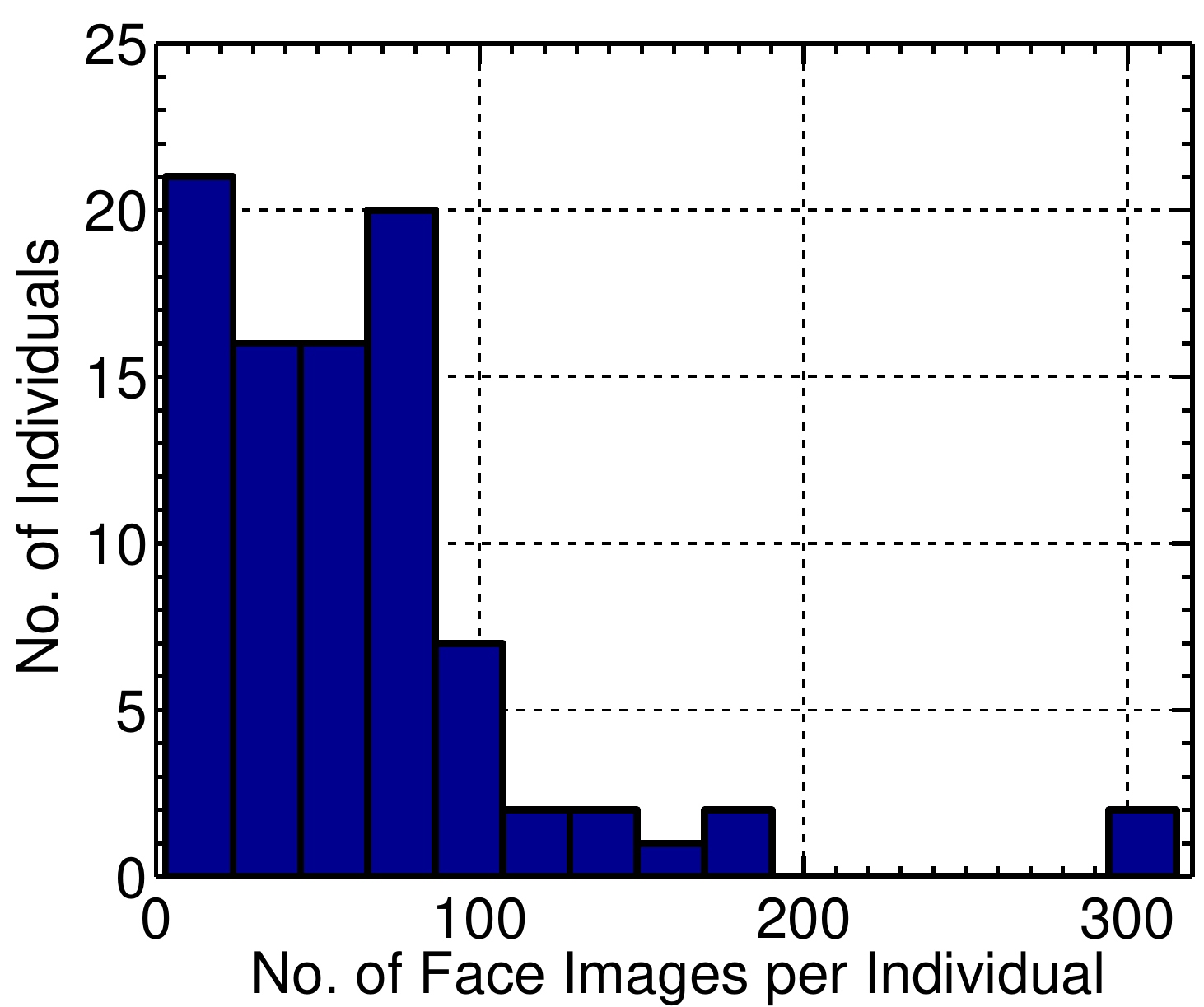}\label{fig:chimp_imgsPerInd}}
  \caption{Histograms of the number of face images per (a) lemur, (b) golden monkey, and (c) chimpanzee. The total number of distinct lemurs, golden monkeys, and chimpanzees in LemurFace, GoldenMonkeyFace, and ChimpFace datasets are 129, 49, and 90, respectively.}
\end{figure}

The LemurFace dataset contains 3,000 face images of 129 lemur individuals from 12 different species (Figure~\ref{fig:lemur_species}) which were photographed by one of the authors at the Duke Lemur Center in North Carolina. Images of lemurs were acquired using an 8 megapixel camera on a mid-range smartphone device, the LG Nexus 5\footnote{\url{https://www.gsmarena.com/lg_nexus_5-5705.php}}. Lemurs were labeled according to the names given to them by the Duke Lemur Center (\eg Alena, Ma'at, West). The eye and chin locations of lemurs were manually annotated by us and any image where both of the lemur's eyes were not clearly visible is removed from the dataset, resulting in a total of 3,000 images. In addition, to account for variation in environmental conditions, we acquired images of the same lemur on two consecutive days. Each individual is photographed both indoors and outdoors. A histogram of the number of images per lemur individual is shown in Figure~\ref{fig:lemur_imgsPerInd}.

\begin{table*}[!t] 
\caption{Summary of LemurFace, GoldenMonkeyFace, and ChimpFace datasets.}
\centering
\begin{tabular}{lcccc}
\noalign{\hrule height 1.5pt}
 & LemurFace & GoldenMonkeyFace & ChimpFace\\
 \hline
 Number of Images & 3,000 & 1,450 & 5,559\\
 Number of Individuals & 129 & 49  & 90\\
 Number of images/individual & [7, 42] & [2, 120] & [3, 315]\\
 Average number of images/individual & 23 & 30 & 63 \\
  \hline
\noalign{\hrule height 1.5pt}
\end{tabular}
\label{table_dataset}
\end{table*}

\subsection{GoldenMonkeyFace Dataset}
Our GoldenMonkeyFace dataset consists of 1,450 face images of 49 golden monkeys (\textit{Cercopithecus mitis kandti}). A total of 241 short video clips (average duration of 6 seconds) were shot by one of the authors using a Nikon Coolpix B700\footnote{\url{https://www.nikonusa.com/en/nikon-products/product/compact-digital-cameras/coolpix-b700.html}} at the Volcanoes National Park in Rwanda. Image frames were extracted from each of the video clips and were then cropped and aligned as described in section~\ref{sec:alignment}. Figures~\ref{fig:gm_samples} and~\ref{fig:gm_imgsPerInd} show example golden monkey face images and a histogram of the number of face images per golden monkey, respectively.

\subsection{ChimpanzeeFace Dataset}
Loos and Ernst provided two chimpanzee face datasets, \textit{C-Zoo} and \textit{C-Tai}, which were extended by Freytag~\etal~\cite{loos},~\cite{freytag}. The \textit{C-Zoo} dataset is comprised of 2,109 face images of 24 chimpanzees and 5,078 face images of 78 individuals are from the \textit{C-Tai} dataset. Eye and mouth center locations are manually annotated for all the images by domain experts. 

Due to the small number of individuals present in the \textit{C-Zoo} dataset, we merged \textit{C-Zoo} and \textit{C-Tai} datasets to form \textit{ChimpFace} and removed all individuals that have less than 3 face images. In total, we have 5,559 images of 90 chimpanzees. Figures~\ref{fig:chimp_samples} and~\ref{fig:chimp_imgsPerInd} show face images of a few chimpanzees from the \textit{ChimpFace} dataset and a histogram of the number of face images per chimpanzee, respectively.

\section{Methodology}
\label{sec:method}
In this section, we introduce the proposed system for aligning and matching the primate face photos. Then, we report experiments to evaluate our system and compare it with existing methods in Section~\ref{experiments}.

\begin{figure}[!t]
  \centering
   \captionsetup[subfigure]{labelformat=empty}
  \subfloat[Original]{\includegraphics[height=0.8in]{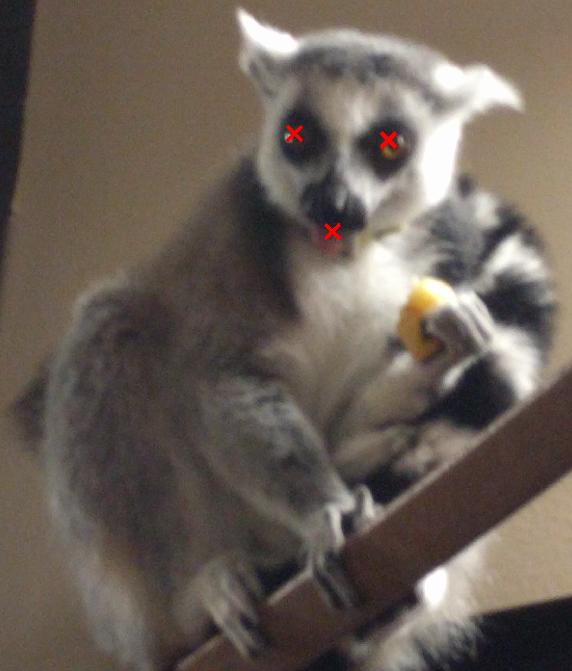}}\hfil
  \subfloat[Aligned]{\includegraphics[height=0.8in]{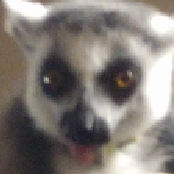}}\vfil
  \subfloat[Original]{\includegraphics[height=0.8in]{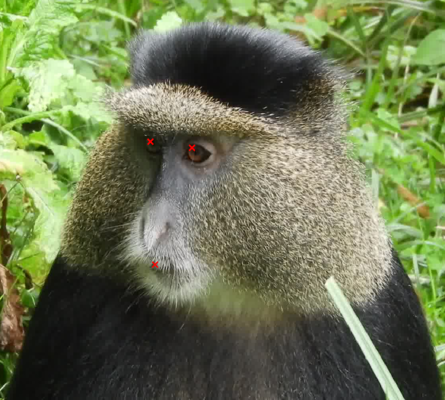}}\hfil
  \subfloat[Aligned]{\includegraphics[height=0.8in]{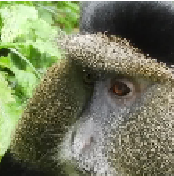}}\vfil
  \subfloat[Original]{\includegraphics[height=0.8in]{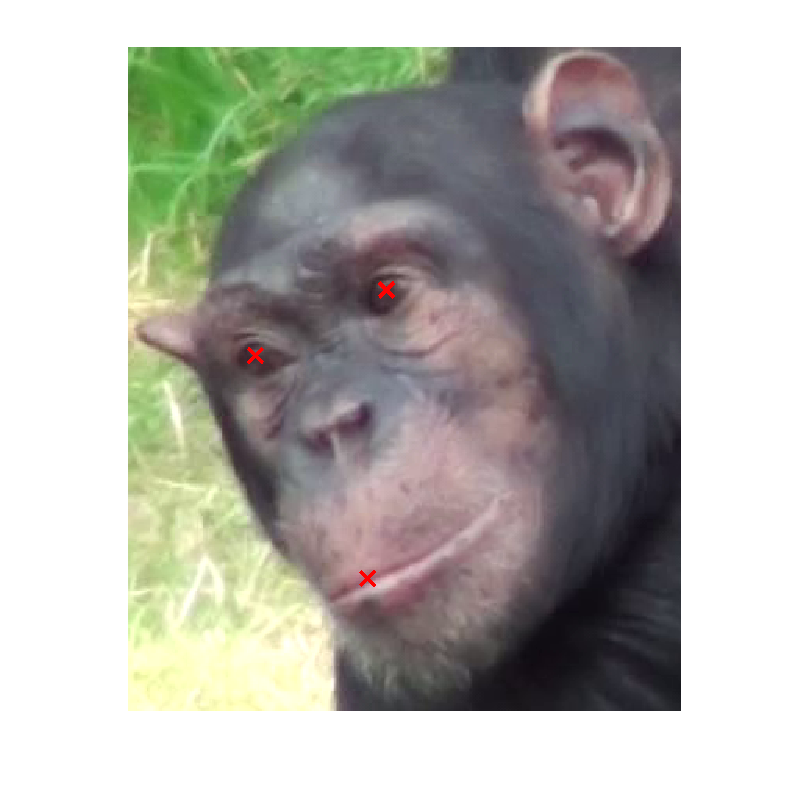}}\hfil
  \subfloat[Aligned]{\includegraphics[height=0.8in]{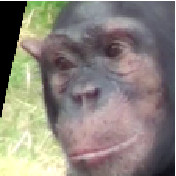}}
  \caption{Primate face images are aligned using a similarity transform.}
  \label{fig:alignment}
\end{figure}

\begin{figure*}[!t]
  \centering
  \includegraphics[width=0.7\linewidth]{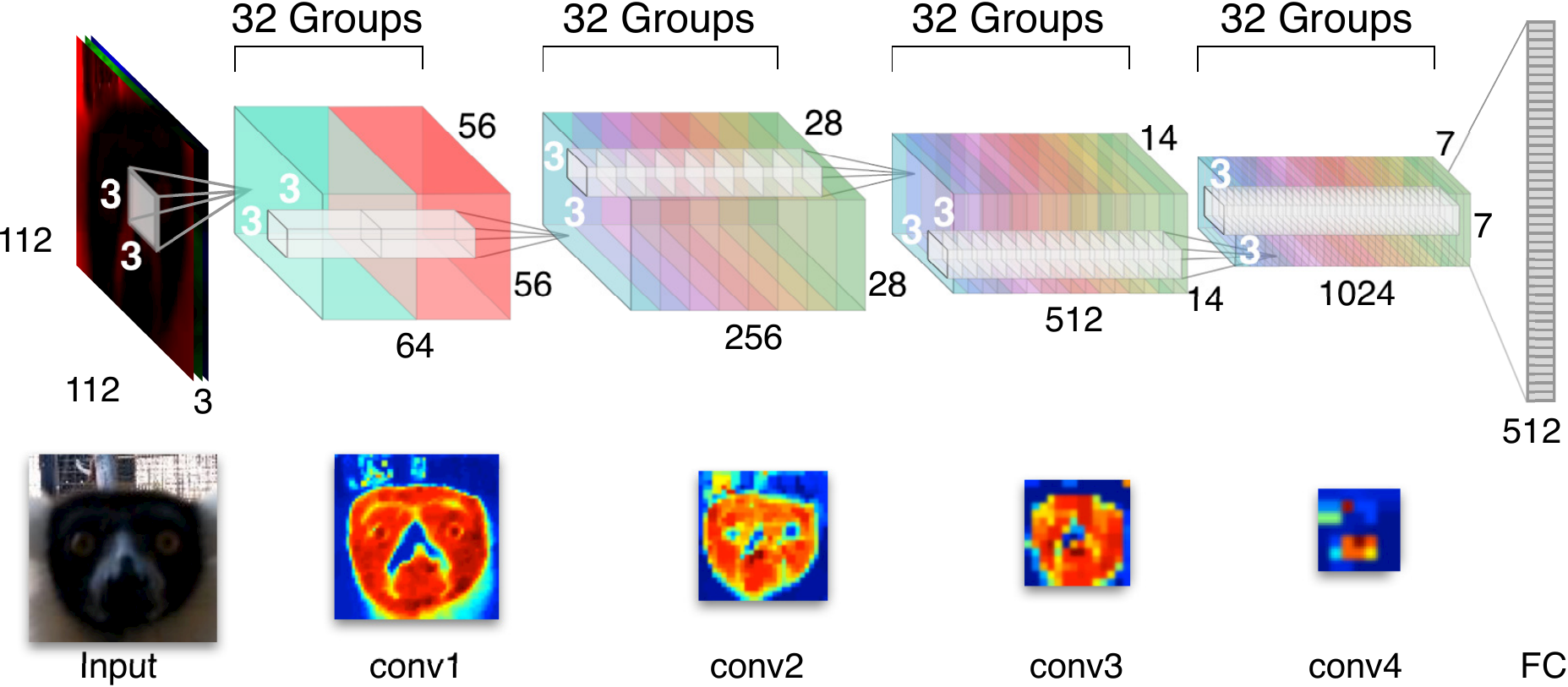}
  \caption{Proposed PrimNet Architecture. A heat map of the intermediate representation of the input lemur's face is shown below each intermediate layer of the network.}
   \label{fig:arch}
\end{figure*}

\subsection{Face Alignment}
\label{sec:alignment}
The primary challenge in designing face recognition systems for primates is to first detect and then align the face images. Due to a lack of large primate face datasets  of the three endangered species considered here, training a face detector specifically for them is not feasible. Face detection also comes with some additional challenges due to the presence of variations in hair and fur, low contrast between eyes and background, and variation in eye colors across different individuals. For these reasons, all the face images in our experiments are manually annotated with three landmarks, namely left eye, right eye and mouth center. These landmarks are used to construct a ``landmark" template using the following procedure. 

Let $[x_{ij}, y_{ij}]^T$ be the landmark locations for the $i^{th}$ image in the dataset, where left eye, right eye, and mouth center coordinates are denoted as $(x_{i1}, y_{i1})$, $(x_{i2}, y_{i2})$, and $(x_{i3}, y_{i3})$. We generate a 6-element vector $L_i = [ \widetilde{x}_{i1}, \widetilde{x}_{i2}, \widetilde{x}_{i3}, \widetilde{y}_{i1}, \widetilde{y}_{i2}, \widetilde{y}_{i3}]$, where $\widetilde{x}_{ij} = \left(x_{ij} - \frac{1}{3}\sum_{k=1}^{3}x_{ik}\right)$ and similarily for $\widetilde{y}_{ij}$.

Then, we compute the ``landmark template'' for a dataset of $N$ images by
\begin{align*}
    t = \frac{1}{N} \sum_{i=1}^{N} \frac{L_i}{||L_i||_2^2}.
\end{align*}

We represent a similarity transform by
\[
    \begin{bmatrix}
     t_x \\ 
     t_y
    \end{bmatrix}
    = 
    \begin{bmatrix}
     s*cos(\theta) & -s*sin(\theta) \\
     s* sin(\theta) & s*cos(\theta)
    \end{bmatrix}
    \begin{bmatrix}
        x \\
        y
    \end{bmatrix}
    + 
    \begin{bmatrix}
        m_x \\
        m_y
    \end{bmatrix},
\]

 where $s$, $\theta$, and $(m_x, m_y)$ are the scale, rotation, and translation parameters, respectively. To solve for the parameters, we rewrite the above as a system of linear equations, $Ax=b$. To solve for the parameters, we obtain a least squares estimate through, $x = (A^TA)^{-1}A^Tb$, where $x = [s * cos(\theta), s*sin(\theta), m_x, m_y]^T$. Figure~\ref{fig:alignment} outlines the methodology for aligning primate face images. In a real-life setting, the user is expected to only manually annotate the three landmarks before submitting it to PrimNet for recognition.

\subsection{PrimNet}
To learn robust face representations for primates, we developed a new CNN architecture, which we call \textit{PrimNet}. One of the requirements of deep neural networks is a sufficiently large dataset to learn numerous network parameters. For human faces, data of this scale is easy to obtain. For other primates, especially the endangered ones, the availability of face datasets is limited.  We found that SphereNet-4~\cite{liu2017sphereface}, one of the smaller face recognition networks, suffers from overfitting when trained on our primate datasets. Hence, we introduced two modifications to the \textit{SphereNet-4} architecture in designing the \textit{PrimNet}:
\begin{itemize}
    \item Reduced the number of parameters by making the network \textit{sparser} through the group convolution stratagem for all the layers~\cite{xie2017aggregated}, followed by channel shuffling~\cite{zhang2017shufflenet}.
    \item Enhanced the discrimination power of hidden layers by making the network \textit{wider} via increased number of channels.
\end{itemize}

In a traditional CNN architecture, each convolution filter applies to all the channels in the input feature map. But in group convolution, as in ShuffleNet~\cite{zhang2017shufflenet}, each convolution filter only applies to a subset of the input channels, thereby significantly reducing the number of parameters. It is important to note that if all the layers adopt group convolution, then the information in each group is isolated and never exchanged. Group shuffling operation after the convolution was proposed to handle this~\cite{zhang2017shufflenet}. Through grouping and shuffling for the four convolution layers, PrimNet becomes a sparse network, with a total of only $9.92\times10^5$ parameters. In comparison, Sphere-4 has $1.26\times 10^7$ parameters and ShuffleNet has around $1.4\times10^8$ parameters. Reducing the number of filters limits the dimensionality of the intermediate layers, however, increasing the sparsity does not inhibit their representation power. Figure~\ref{fig:arch} illustrates the proposed network architecture. PrimNet is trained using the \textit{AM-Softmax function}, which has been shown to be effective in learning human face representations~\cite{amsoftmax}.



\section{Experiments}
\label{experiments}
We evaluate the performance of PrimNet on three tasks: (i) verification, (ii) closed-set identification, and (iii) open-set identification. For each experiment, we evaluate the performance of primate individualization models using 5-fold cross-validation.

In our study, genuine comparisons are formed by choosing one face image from each primate individual's imagery as the query image and comparing it to the same individual's \textit{template}, \ie, all remaining face images of the individual. We repeat the query and template split until each image from each individual has been used as a query image. In a similar fashion, we form impostor comparisons by considering a query image of a primate individual and comparing it to the all other individuals' templates. For both genuine and impostor comparisons, a similarity score is obtained by computing the cosine similarity between the corresponding feature vector. The highest similarity score within a template acts as the individual's overall similarity score. In practical usage, the verification scenario is invaluable for gathering evidence of live primate trafficking. Suppose a photograph of a certain primate appears illegally for sale on a social media account, and a similar photo appears on a different account. The verification task can assist in confirming whether the two photographs belong to the same individual. Confirming an individual's identity through verification can greatly aid in closing the loop on online primate trafficking by illuminating the network of smugglers and traders involved.

Verification accuracy is reported as the mean and standard deviation of True Accept Rates (TARs) at 1\% and 0.1\% False Accept Rates (FARs) across the 5 folds.

\textit{Identification} (1:N search) searches a dataset (gallery) to determine the identity of an individual from a given probe (query) image. In \textit{closed-set identification}, the probe individual is assumed to be enrolled in the gallery. Through closed-set identification, missing individuals can be identified and returned to the colony. In our experiments, closed-set identification is conducted by randomly choosing a face image from each primate individual as the probe image and the rest of the individual's imagery are kept in the gallery. As in verification, the probe image is compared to each image within an individual's template, and the highest similarity score from these comparisons is the individual's overall similarity score. Then, the individual with the highest similarity score is considered to be the probe's true mate in the gallery. The Cumulative Accuracy is computed as the fraction of correctly identified (retrieved) individuals at Rank 1. In the \textit{open-set} identification, the individual in the query image may not be previously enrolled in the gallery and thus, the recognition system must be capable of indicating that the individual in the probe is not in the dataset. For open-set experiments, we extend the probe set by incorporating primate face images of individuals not present in the gallery. Detection and Identification Rate (DIR) at 1\% FAR and Rank 1 retrieval accuracy is reported. In both closed-set and open-set identification scenarios, for each of the 5 folds, we run 100 trials of randomly splitting the test set into probe and gallery sets.

\subsection{Baseline}
To obtain a baseline performance, we evaluate the individualization accuracy of LemurFaceID~\cite{crouse} which is based on Local Binary Patterns (LBP) features~\cite{lbp}. Using a training set of 104 lemurs and a testing set of 25 lemurs, we achieve a baseline verification performance of 81.90\% $\pm$ 3.69\% TAR at 1\% FAR and 90.82\% $\pm$ 1.80\% closed-set identification accuracy at Rank-1 across the five folds. Table~\ref{tab:results} summarizes the results.

\begin{table*}[t]
\centering
\caption{Performance on three different primates: Lemurs, Golden Monkeys, and Chimpanzees.}
\resizebox{\linewidth}{!}{
\begin{tabular}{|l||c|c|c||c|c|c||c|c|c|}
\noalign{\hrule height 1.5pt}
                & \multicolumn{3}{c||}{\textbf{Lemurs}}
                & \multicolumn{3}{c||}{\textbf{Golden Monkeys}}                    & \multicolumn{3}{c|}{\textbf{Chimpanzees}} \\ \hline
\textbf{Method} & \textbf{Verification} & \textbf{Closed-set} & \textbf{Open-set} & \textbf{Verification} & \textbf{Closed-set} & \textbf{Open-set} & \textbf{Verification} & \textbf{Closed-set} & \textbf{Open-set} \\ 
\noalign{\hrule height 1.5pt}
                & 1\% FAR               & Rank-1              & Rank-1            & 1\% FAR               & Rank-1              & Rank-1            & 1\% FAR               & Rank-1       & Rank-1           \\
\noalign{\hrule height 1.2pt}
Baseline~\cite{crouse}  & 81.90 $\pm$ 3.69 & 90.82 $\pm$ 1.80 & N/A & 74.88 $\pm$ 6.75 & 89.33 $\pm$ 7.68 & N/A& 44.62 $\pm$ 4.38 & 70.16 $\pm$ 3.36 & N/A\\ \hline
SphereFace-20~\cite{liu2017sphereface}  & 79.40 $\pm$ 5.82  & 92.45 $\pm$1.67 & 80.83 $\pm$ 4.48 &65.18 $\pm$ 12.28         & 87.32 $\pm$ 4.57        &      61.15 $\pm$ 12.80              & 48.62 $\pm$ 6.23          & 75.49 $\pm$ 3.80 &  30.75 $\pm$ 12.41            \\ \hline
SphereFace-4~\cite{liu2017sphereface}    & 73.6 $\pm$ 5.81                 & 90.18 $\pm$ 1.37                       & 72.29 $\pm$ 9.49 & 72.53 $\pm$ 6.57          & 87.49 $\pm$ 3.77        &  69.43 $\pm$ 9.27                & 53.92 $\pm$ 2.57          & 74.19 $\pm$ 3.74        & 35.85 $\pm$ 8.22  \\ \hline
FaceNet~\cite{facenet}     & 55.52 $\pm$ 7.88                & 87.06 $\pm$ 9.63                       & 56.12 $\pm$ 1.93    & 50.12 $\pm$ 15.31         & 73.47 $\pm$ 8.81  &  49.69 $\pm$ 9.54         & 17.89 $\pm$ 7.93  & 59.75 $\pm$ 8.64  &   4.86 $\pm$ 3.38        \\ \hline
PrimNet          & \textbf{83.11 $\pm$ 5.31}                & \textbf{93.76 $\pm$ 0.90}                       & \textbf{81.73 $\pm$ 2.36} & \textbf{78.72 $\pm$ 5.80}          & \textbf{90.36 $\pm$ 0.92}        &     \textbf{66.11 $\pm$ 7.99}              & \textbf{59.87 $\pm$ 3.34}          & \textbf{75.82 $\pm$ 1.25}        &        \textbf{37.08 $\pm$ 11.22}                  \\
\noalign{\hrule height 1.5pt}
\end{tabular}}
\label{tab:results}
\end{table*}

\begin{table}[!t]
\centering
\caption{Inference speed and model size of different networks.}
\label{usability}
\resizebox{\linewidth}{!}{
\begin{tabular}{|l|c|c|}
\noalign{\hrule height 1.5pt}
{\textbf{Method}} & \textbf{Inference Speed (ms / img)} & \textbf{Model Size (MB)} \\ 
\noalign{\hrule height 1.5pt}
SphereFace-20~\cite{liu2017sphereface}                       & 17.26                                 & 87                       \\ \hline
SphereFace-4~\cite{liu2017sphereface}                       & \textbf{13.05}                                 & 48                       \\ \hline
FaceNet~\cite{facenet}                             & 40.42                                 & 90                       \\ \hline
\textbf{PrimNet}                             & 23.58                                 & \textbf{3.9}       \\
\noalign{\hrule height 1.5pt}              
\end{tabular}}
\end{table}

\begin{figure}[!t]
  \centering
  \includegraphics[width=0.8\linewidth]{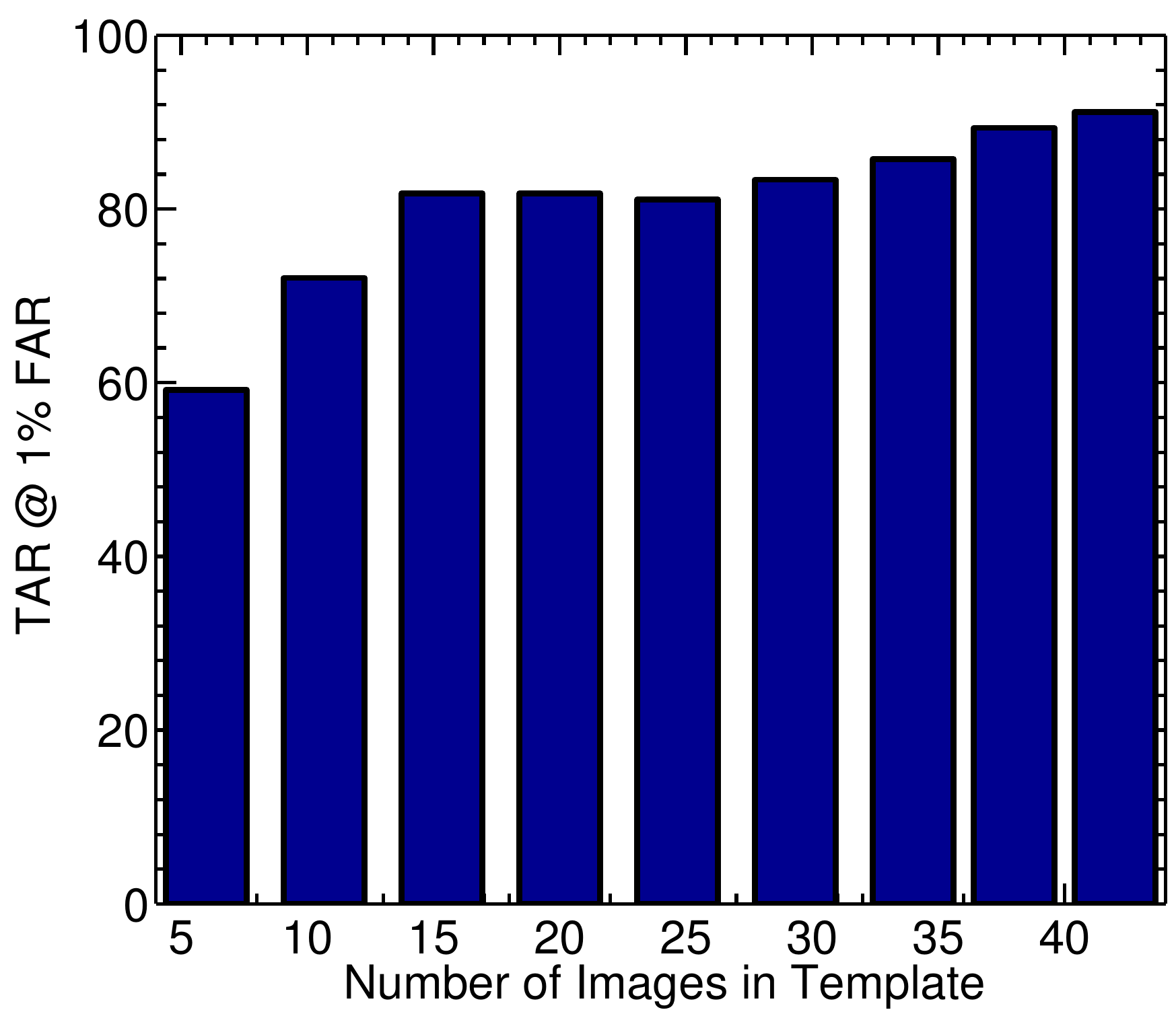}
  \caption{Verification accuracy with respect to varying number of images per template. Performance increases with an increased number of images in a template.}
  \label{verification_vary}
\end{figure}

\subsection{Human FR to Primate FR}
\label{sec:finetuning}
Since we have related the primate face recognition problem to human face recognition, one might wonder whether CNNs trained for human faces are also suited for the primate faces. We evaluate the performance of SphereFace and FaceNet on \textit{LemurFace}, \textit{GoldenMonkeyFace}, and \textit{ChimpFace} datasets by finetuning the two state-of-the-art human face recognition networks. We use pre-trained network parameters for SphereFace and FaceNet\footnote{SphereFace is trained on 494,414 face images of 10,575 humans (CASIA WebFace~\cite{casia}) and FaceNet is trained on 3.31 million face images of 9,131 humans (VGGFace2~\cite{vgg}).} as initialization for the lemur individualization task. For SphereFace, we use 20 and 4 hidden layer models, denoted as \textit{SphereFace-20} and \textit{SphereFace-4}, respectively.

To illustrate this idea, we show the performance of fine tuned SphereNet and FaceNet on lemur face data. For each of the five folds, 104 lemurs are used for training and the remaining 25 are kept for testing. In the verification scenario, there are 625 genuine comparison scores and 15,625 impostor comparison scores in each fold. For open-set identification, we extend the probe set by including 953 images of 449 lemur individuals downloaded from the internet. Table~\ref{tab:results} reports the evaluation results on LemurFace. We conclude that even though human face recognition systems can be finetuned for use with lemurs, achieving acceptable face recognition performance for primates in the wild requires further enhancement.

\begin{figure*}[!t]
  \centering
   \captionsetup[subfigure]{labelformat=empty}
  \begin{minipage}{\textwidth}
  \subfloat[Lemurs]{\includegraphics[width=0.3\linewidth]{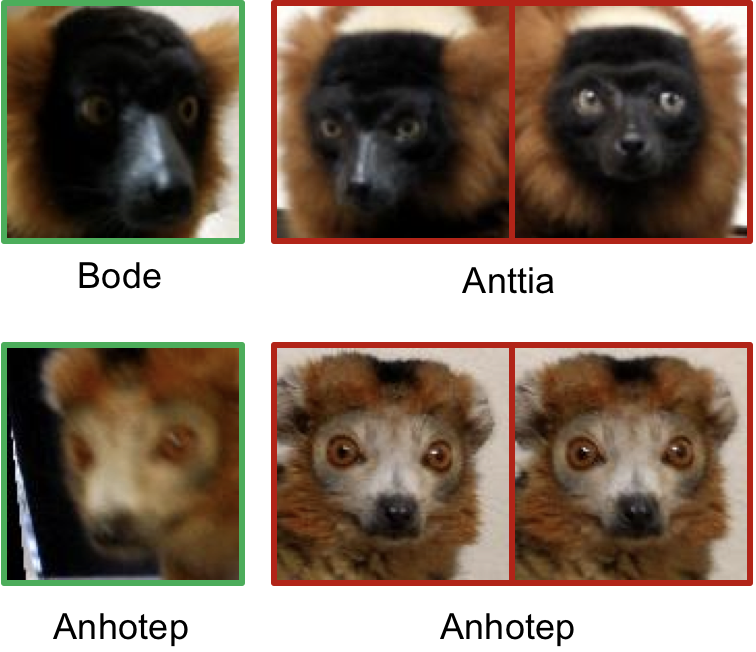}}\hfil
  \subfloat[Golden Monkeys]{\includegraphics[width=0.3\linewidth]{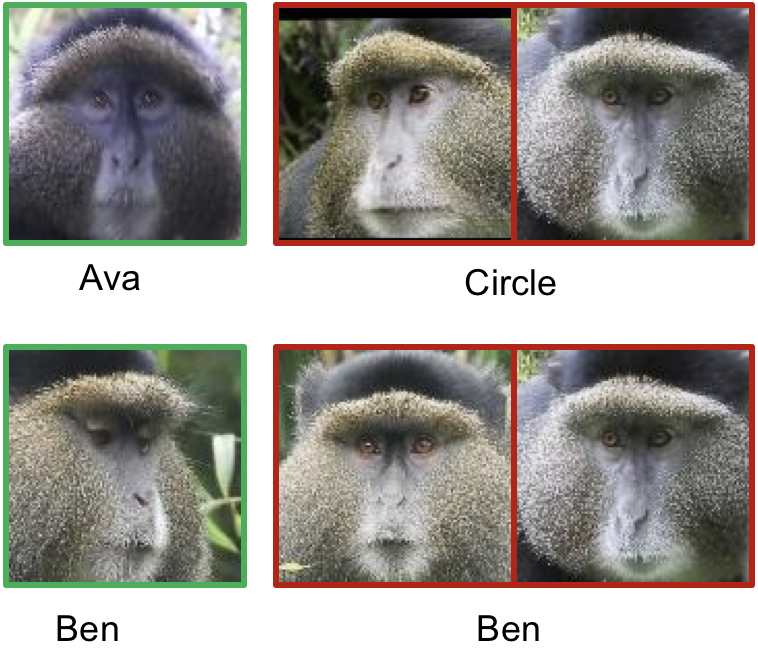}}\hfil
  \subfloat[Chimpanzees]{\includegraphics[width=0.3\linewidth]{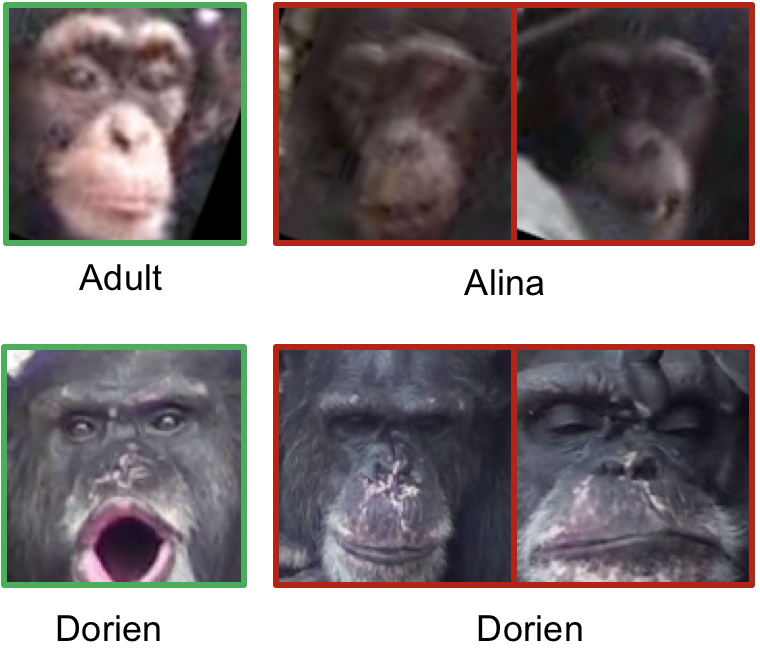}}
  \end{minipage}
  \label{verification_failure}
  \caption{Example cases where PrimNet fails to verify primate individuals.~\textit{Top row:}~Two distinct individuals that are \textit{falsely accepted} at 1\% FAR.~\textit{Bottom row:}~Same individuals that are \textit{falsely rejected} at 1\% FAR. Green box denotes the probe image and two images from the template are shown within a red box. These errors are caused primarily due to poor quality of the query, change in expression and viewpoint.}
\end{figure*}

\begin{figure}[!t]
  \centering
  \subfloat[Species Selection]{\includegraphics[width=0.32\linewidth]{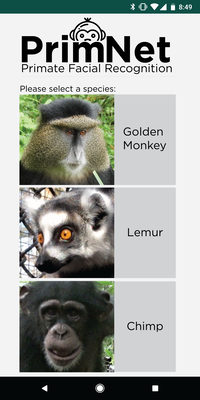}\label{fig:choice}}\hfil
  \subfloat[Verification]{\includegraphics[width=0.32\linewidth]{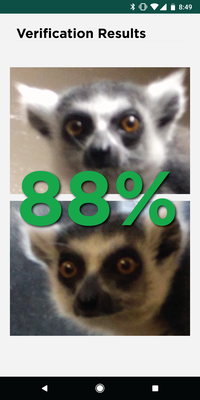}\label{fig:veri_id}}\hfil
  \subfloat[Identification]{\includegraphics[width=0.32\linewidth]{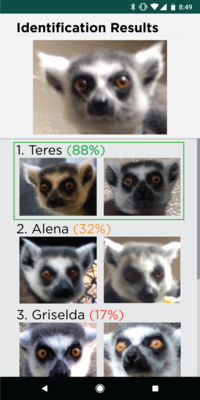}\label{fig:gallery}}
  \label{android_app}
  \caption{Screenshots from the PrimNet Android application.}
\end{figure}

\subsection{PrimNet: Lemurs}
The PrimNet architecture is trained on LemurFace dataset from scratch. Table~\ref{tab:results} summarizes the results. Performance of PrimNet is superior to those of baseline networks: LemurFaceID, SphereFace, and FaceNet.

To understand the variation in verification performance across the five folds, we plot the TAR at 1\% FAR with respect to varying number of images in the template. As  expected, Figure~\ref{verification_vary} shows that as the number of images in the template increases, the verification accuracy improves. For reliable verification, it is recommended to keep at least 15 images in a lemur's template. It is important to note that we currently use a single probe image during verification. Indeed, increasing the number of probe images for verification can further enhance the verification performance.  

\subsection{PrimNet: Golden Monkeys}
We used 39 individuals for training and the remaining 10 individuals for testing. In each fold, we have approximately 280 genuine and 2,520 impostor comparison scores. For each of the 100 trials, we have 10 probe images and 270 gallery images in the gallery, across the five folds, for closed-set identification performance. See Table~\ref{tab:results} on the performance of PrimNet and other networks on the GoldenMonkeyFace dataset.

\subsection{PrimNet: Chimpanzees}
Using 5-fold cross-validation, training and testing datasets for ChimpFaces consists of 72 and 18 chimpanzees, respectively. For each fold, we compute 1,259 genuine and 21,403 impostor scores. For closed-set identification, we have 18 chimp face images in the probe set and around 1,241 gallery images, across the five folds. We find that PrimNet outperforms other networks in Table~\ref{tab:results}.

Figure~\ref{verification_failure} shows examples of the failure cases, which are primarily caused due to poor quality probe image. Extreme variations in an individual's pose can adversely affect the verification performance. From Table~\ref{usability}, we find that PrimNet achieves inference speed comparable to other state-of-the-art networks (24 ms per image) while maintaining high accuracy. In addition, the greater advantage of PrimNet is in its size. With a mere storage space requirement of 3.9 MB, PrimNet is well suited for deployment on embedded systems such as smartphones.

\section{Mobile App}
\label{implementation} 
We developed an Android mobile application which can be used for primate individualization in the wild\footnote{The application source code can be found at~\url{https://github.com/ronny3050/PrimateFaceRecognitionAndroid}.}. We trained the PrimNet architecture on the entire LemurFace, GoldenMonkeyFace and ChimpFace datasets. Currently, the app offers the user a choice to individualize one of the three primates (See Figure~\ref{fig:choice}). On choosing the primate of interest, the app loads the gallery of individuals currently in the dataset. The user may wish to either (i) verify whether a set of images belong to the same individual, or (ii) identify the individual in a given probe image by searching the gallery. In identification mode, the top three ranks from the possible candidates list are displayed to the user with the associated similarity scores. In verification mode, results are given by the similarity score between the query and template. Screenshots for verification and identification modes are shown in Figures~\ref{fig:veri_id} and~\ref{fig:gallery}. 

\section{Conclusion}
\label{conclusion}
We have designed a new primate face recognition network, PrimNet, using a convolutional neural network (CNN) architecture. We compared the performance of PrimNet to a benchmark primate recognition system, LemurFaceID, as well as two open-source human face recognition systems, SphereFace~\cite{liu2017sphereface} and FaceNet~\cite{facenet}. We evaluated the systems on three primate datasets: \textit{LemurFace}, \textit{GoldenMonkeyFace}, and \textit{ChimpFace}. The performance of PrimNet was superior to the other networks in both verification (1:1 comparison) and identification (1:N search) scenarios. 

%

As primate species are threatened by habitat loss, hunting, and trafficking, it is imperative that primate researchers and conservationists have efficient and effective tools to reliably and safely monitor these animals. We believe the PrimNet primate face recognition system can greatly aid in these efforts to ensure that these endangered animals are protected. Through our collaborations with domain experts and field researchers, we plan to enlarge our primate datasets to further improve the recognition accuracy and to even develop a primate face detector. In addition, we also plan on evaluating PrimNet on datasets comprising of other endangered primate species.

\section{Acknowledgement}
The authors would like to express their gratitude to Duke Lemur Center for their assistance in \textit{LemurFace} dataset acquisition\footnote{~\url{http://lemur.duke.edu/}}. We also acknowledge Daniel Stiles\footnote{~\url{https://freetheapes.org/}} and Dr. Alison Fletcher for their guidance and support. In addition, we thank Dian Fossey Gorilla Fund International~\footnote{\url{https://gorillafund.org/}} and Rwanda Development Board~\footnote{\url{http://rdb.rw/}} for their support on the work with golden monkeys.		

{\small
\bibliographystyle{unsrt}
\bibliography{submission_example}
}

\end{document}